\documentclass[10pt,twocolumn,letterpaper]{article}

\usepackage{wacv}              

\usepackage{graphicx}
\usepackage{amsmath}
\usepackage{amssymb}
\usepackage{booktabs}
\usepackage{mathtools}
\usepackage{multirow}
\usepackage{color, colortbl}
\usepackage{diagbox}
\usepackage{soul}
\usepackage{xcolor}
\usepackage{enumitem}

\usepackage[pagebackref,breaklinks,colorlinks]{hyperref}

\newcommand{\ourmodel}{GPG2A}
\newcommand{\+}{v2}

\usepackage[capitalize]{cleveref}
\crefname{section}{Sec.}{Secs.}
\Crefname{section}{Section}{Sections}
\Crefname{table}{Table}{Tables}
\crefname{table}{Tab.}{Tabs.}

\def\wacvPaperID{980} 
\def\confName{WACV}
\def\confYear{2025}

\newcommand{\supp}[1]{\textcolor{gray}{\textbf{\textit{#1}}}}

\begin{document}


\title{Cross-View Meets Diffusion: Aerial Image Synthesis with Geometry and Text Guidance}

\author{Ahmad Arrabi$^{1\dag}$, Xiaohan Zhang$^{1 \dag}$,  Waqas Sultani$^{2}$, Chen Chen$^{3}$, Safwan Wshah$^{1*}$ \\
$^1$ Vermont Artificial Intelligence Lab, Department of Computer Science, University of Vermont \\ $^2$ Intelligent Machines Lab, Information Technology University \\
$^3$Center for Research in Computer Vision, University of Central Florida \\
{\small $^\dag$ These authors contributed equally. $^*$ Corresponding and senior author.} \\
}

\maketitle

\begin{abstract}

Aerial imagery analysis is critical for many research fields. However, obtaining frequent high-quality aerial images is not always accessible due to its high effort and cost requirements. 
One solution is to use the Ground-to-Aerial (G2A) technique to synthesize aerial images from easily collectible ground images. However, G2A is rarely studied, because of its challenges, including but not limited to, the drastic view changes, occlusion, and range of visibility. In this paper, we present a novel  \textbf{G}eometric \textbf{P}reserving Ground-to-Aerial (\textbf{G2A}) image synthesis (\ourmodel{}) model that can generate realistic aerial images from ground images. \ourmodel{} consists of two stages. The first stage predicts the Bird's Eye View (BEV) segmentation (referred to as the BEV layout map) from the ground image. The second stage synthesizes the aerial image from the predicted BEV layout map and text descriptions of the ground image. To train our model, we present a new multi-modal cross-view dataset, namely VIGOR\+, built upon VIGOR~\cite{Vigor} with newly collected aerial images, maps, and text descriptions.
Our extensive experiments illustrate that \ourmodel{} synthesizes better geometry-preserved aerial images than existing models. We also present two applications, data augmentation for cross-view geo-localization and sketch-based region search, to further verify the effectiveness of our \ourmodel{}. The code and dataset are available at \url{https://gitlab.com/vail-uvm/gpg2a}.
\end{abstract}

\section{Introduction}
\label{sec:intro}

\begin{figure}[!ht]
    \centering\includegraphics[page=14,width=0.47\textwidth, trim={24cm 24.3cm 32.1cm 16.2cm}, clip]{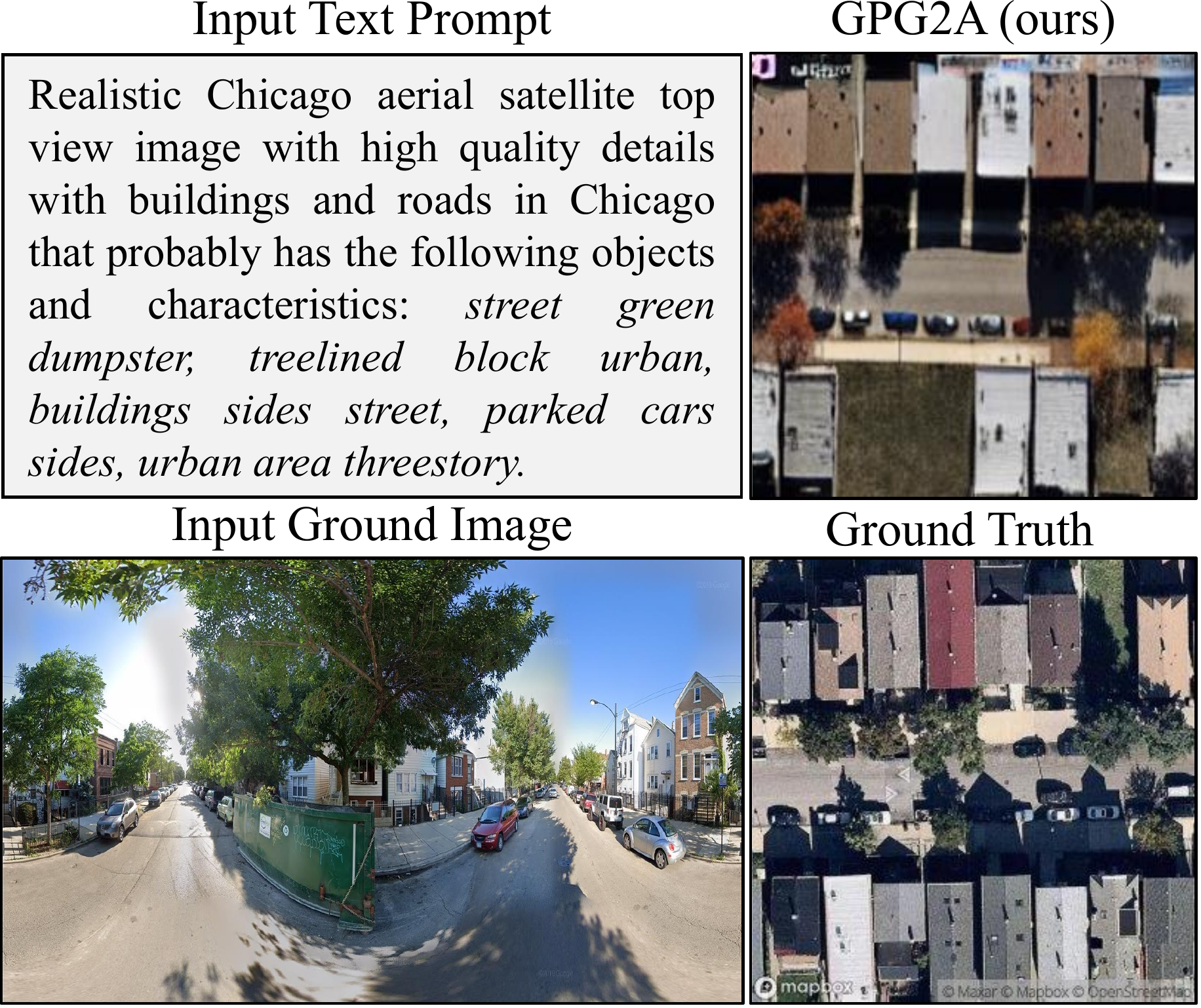}
    \caption{An example generated aerial image (top right) by our \ourmodel{} from the input text prompt (top left) and the ground image (bottom left). The ground truth aerial image is on the bottom right.}
    \label{fig:teaser}
    \vspace{-15pt}
\end{figure}

Unlike satellite images, which are low in resolution and can be obscured by clouds~\cite{sentinel,sat1,landsat}, aerial images capture more detailed views, benefiting various applications, such as land use classification~\cite{land_use1,land_use2}, urban planning~\cite{urban_planning}, transportation~\cite{transportation1,transportation2,transportation3}, socioeconomic studies~\cite{socioeco,econmics_ieee}, and cross-view geo-localization (CVGL)~\cite{geodtr,geodtr+,Vigor,SAFA,CVUSA}. However, current aerial images are limited by the high effort and cost required to capture them, as they are often captured by Unmanned Aerial Vehicles (UAVs) or drones. For example, New York State's government annually captures aerial images for only one-third of its counties~\cite{NY_aerial}. Security concerns also restrict drone use at low altitudes in urban areas, limiting applications and preventing frequent updates. These accessibility challenges are more common in developing countries. In contrast, ground images are far more available and cost-effective, especially in the recent advanced cars and autonomous vehicles. Also, crowdsourcing platforms like Mapillary~\cite{mapillary} see tons of daily uploads of street-view images. Thus, a promising solution for such challenges is ground-to-aerial (G2A) image synthesis, which aims to generate more frequent aerial images from their corresponding ground views.

Despite the potential of G2A image synthesis, to the best of our knowledge, there has been limited research addressing this task due to its challenges. These challenges include the drastic viewing angle change, object occlusions, and different ranges of visibility between aerial and ground views. Some prior works attempted G2A synthesis mainly leveraging Generative Adversarial Networks (GANs)~\cite{GAN} but lacked explicit geometric constraints~\cite{regmi2018cross} 
or depended on strong priors like segmentation maps of the aerial view~\cite{selectionGAN}. 

In this work, we propose \textbf{G}eometric \textbf{P}reserving Ground-to-Aerial (\textbf{G2A}) image synthesis (\ourmodel{}) model which features a novel two-stage process. The first stage transforms the input ground image into a Bird's Eye View (BEV) layout map. The second stage leverages pre-trained diffusion models~\cite{ddpm,controlnet}, conditioned on the predicted BEV layout map from the first stage, to generate photo-realistic aerial images 
This innovative two-stage pipeline provides three advantages: 1) The problem is simplified by introducing an intermediate BEV layout map stage reducing the domain gap between aerial and ground views. 2) The BEV layout map explicitly preserves the geometry, enhancing the synthesized aerial images by maintaining consistent geometry with ground images and reducing overfitting to low-level details. 3) By leveraging the pre-trained knowledge from diffusion foundation models, our \ourmodel{} can synthesize highly realistic images. 

To further improve the synthesis quality and fuse surrounding information not fully represented in the BEV layout map, such as block types (e.g., commercial or residential), we obtain ground image descriptions from large language models (e.g. Gemini). These descriptions are fed into ControlNet~\cite{controlnet} alongside the BEV layout maps, as shown in~\cref{fig:teaser}. Our research not only addresses G2A synthesis but also proposes the VIGOR\+ dataset, which includes center-aligned aerial-ground image pairs, layout maps, and text descriptions of ground images to train our \ourmodel{}. 

To demonstrate the effectiveness of our \ourmodel{}, we explore its application in two downstream tasks: data augmentation for CVGL and sketch-based region search. We show that synthesized data from our \ourmodel{} can enhance the performance of existing CVGL models. Additionally, we illustrate the potential of synthesized images in sketch-based image retrieval, providing a more explainable and controllable approach. By presenting \ourmodel{}, VIGOR\+, and its applications, we aim to attract more researchers to advance this important and challenging field.

Our contribution can be summarized in three-folds,
\vspace{-7.8pt}
\begin{itemize}[leftmargin=*]
    \item We propose \ourmodel{}, a novel two-stage model that tackles the G2A image synthesis task. The first stage explicitly preserves the geometric layout by predicting the BEV maps from ground images. The second stage synthesizes aerial images by conditioning on the layout maps and text prompts of the ground images by using a diffusion model.\vspace{-7pt}
    \item We put forward a novel multi-modal cross-view dataset, namely, VIGOR\+. Upon the existing VIGOR~\cite{Vigor} dataset, we collected center-aligned aerial images, BEV layout maps, and text descriptions of ground images. VIGOR\+ is the first cross-view dataset with image, text, and map modalities.\vspace{-7pt}
    \item We evaluate our \ourmodel{} by using SOTA CVGL models and a customized FID~\cite{FID} score. Extensive experiments demonstrate the outstanding performance of the proposed \ourmodel{} on both same-area and cross-area protocols of VIGOR\+. Moreover, the proposed approach paves the way for many applications. We demonstrate two downstream applications of our \ourmodel{}: 1) Data augmentation for CVGL and 2) Sketch-based Region Search.
\end{itemize}
\section{Related Work}
\label{sec:related_work}
\noindent\textbf{Cross-View Image Synthesis: }
Regmi \etal~\cite{regmi2018cross} introduced cross-view image synthesis, dividing it into two sub-tasks: Aerial-to-Ground (A2G) and Ground-to-Aerial (G2A) synthesis. A2G synthesizes ground images from aerial images, while G2A tackles the inverse problem. Regmi \etal~\cite{regmi2018cross} tackled these two tasks by conditional GANs~\cite{GAN}. Another GAN-based approach in \cite{selectionGAN} conditions on segmentation maps of the target view, providing strong geometric prior assumptions. 

Recently, the A2G task has been actively studied further by enhancing GANs~\cite{crossview_pano}, with CVGL~\cite{CDE}, and leveraging geometric priors~\cite{geomety_image_syn,shi2022geometry}. However, G2A remains less explored or often simplified by assuming strong priors as conditional inputs. 
This lack of research is attributed to the inherent challenges of the G2A task, such as occlusions and the limited resolution of objects in the ground images.

The most relevant research field to this work is Bird's Eye View (BEV) prediction which aims to predict overhead segmentation from ground views. Most BEV studies are designed for autonomous driving~\cite{BEV1, BEV2, BEV3, BEV4, BEV5, BEV6} focusing on moving objects such as vehicles or pedestrians. In contrast, we aim to predict the view of the BEV image at the pixel space by focusing on static city objects such as buildings, roads, and paths. Therefore, the existing BEV methods are not directly applicable to our work. 

Inspired by the recent success of diffusion models~\cite{ddpm,LDM,controlnet} in various tasks~\cite{Diff_text_to_image,Diff_sound,Diff_video, t2i, pose}
, we propose \ourmodel{} which is a novel two-stage model to solve the G2A image synthesis problem. \ourmodel{} closes the domain gap between the aerial and ground views by introducing an intermediate BEV layout stage. 
Our comprehensive experiments demonstrate that this innovative approach remarkably enhances the quality of the synthesized aerial image.

\noindent\textbf{Cross-View Datasets: }Cross-view geo-localization and cross-view synthesis share many common attributes. Therefore, these two tasks are usually conducted on the same datasets~\cite{CVACT,CVUSA,university1652,Vigor}. However, none of these datasets meet the requirement of our \ourmodel{}, since the absence of corresponding layout maps and text description of ground images. Additionally, some datasets are unsuitable for real scenarios because they lack complex scenes~\cite{wilson2021visual}. For example, CVUSA~\cite{CVUSA} collects images from rural areas in the U.S. CVACT~\cite{CVACT} only contains images from one single city in Australia. The images in University-1652~\cite{university1652} are exclusively for campus buildings.
Fortunately, VIGOR~\cite{Vigor} collected aerial and ground images in four major U.S. cities. However, VIGOR is designed for the many-to-one CVGL task, resulting in the misalignment of the aerial-ground image pairs. This misalignment reduces the co-visibility between the ground and aerial views, making it unsuitable for the G2A task. To this end, we propose VIGOR\+ to accommodate the needs of G2A synthesis. Our proposed dataset will be publicly available for further research.
\section{VIGOR\+ Dataset}
\label{sec:Data}
\begin{figure}[!ht]
    \vspace{-10pt}
    \centering\includegraphics[page=1,width=0.48\textwidth, trim={2.5cm 17.5cm 36.0cm 32.5cm}, clip]{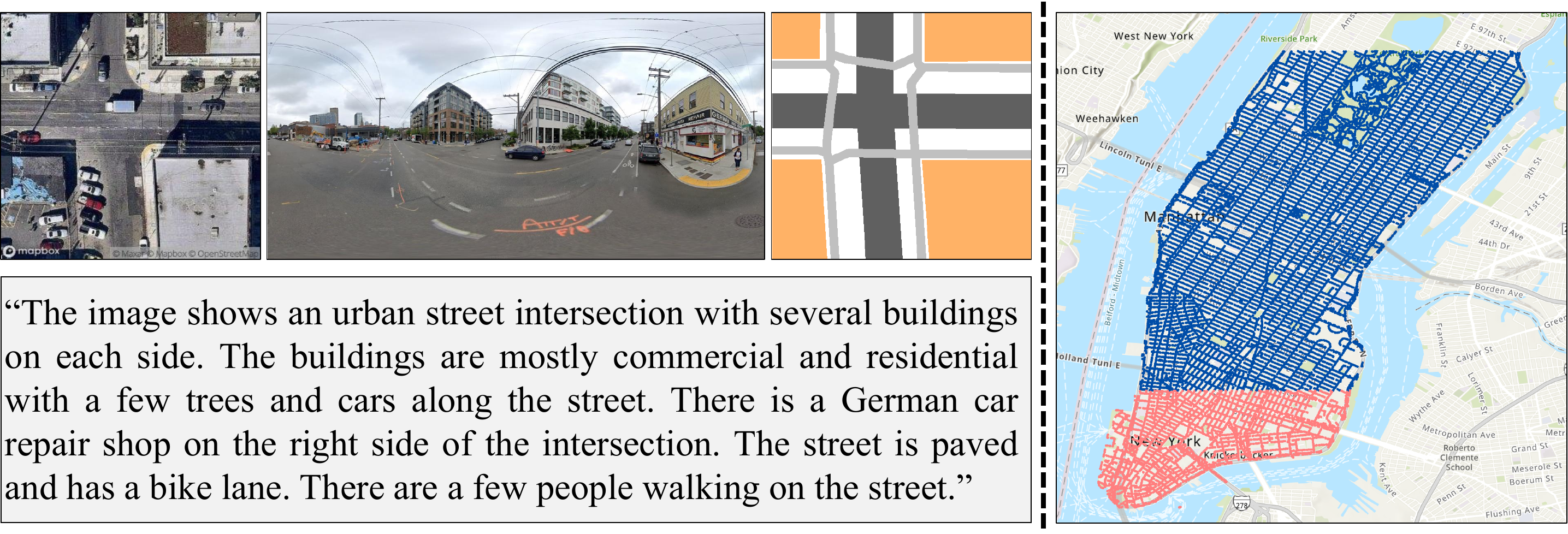}
    \caption{Left: Aerial image (left), ground image (middle), BEV layout map (right), and text description (bottom) from VIGOR\+. Right: The new training (blue lines) and testing (red lines) geographically split of New York City portion of VIGOR\+. The non-overlapping training and testing sets prevent data leakage.}
    \label{fig:dataset_samples}
    \vspace{-10pt}
\end{figure}
As mentioned in~\cref{sec:intro}, we propose VIGOR\+ to accommodate the needs of the G2A image synthesis task. Our solution involves retaining the ground images from VIGOR while re-collecting center-aligned aerial images. In addition to the newly collected aerial images, we enhance the VIGOR dataset by introducing two new modalities: BEV layout maps and text descriptions of ground images. These additional modalities provide rich spatial contextual information and descriptive fine-grained details from the text, resulting in a more robust and comprehensive dataset. Our BEV layout maps offer much more accuracy and contain more classes than previous work~\cite{regmi2018cross} which uses off-the-shelf segmentation models~\cite{lin2017refinenet}.

\noindent\textbf{Aerial Imagery: } For each ground image, we first extract its latitude and longitude and then request an aerial image centered on this location from MapBox~\cite{map_box} API with a resolution of $300 \times 300$ and a zoom level of $18.5$. We empirically chose this zoom level by visually inspecting that the aerial image covers most of the visual areas on ground images. 


\noindent\textbf{BEV Layout Maps: }Accurate BEV Layout Maps are needed to train our \ourmodel{}. Inspired by recent work~\cite{orienternet}, we collect BEV maps through OpenStreetMap~\cite{OpenStreetMap} API with the location of the ground image and a zoom level similar to $18.5$. Specifically, we select $8$ main categories to render with different colors in the BEV layout map, namely, building, parking, playground, forest, water, path, road, and others. The rendered BEV layout map shares the same resolution as the aerial image as of $300 \times 300$.

\noindent\textbf{Text Descriptions: }
Surrounding environment information such as the types of blocks and texture of buildings is valuable in G2A image synthesis. In our \ourmodel{}, the text description is assigned to convey such information to the model. To this end, we utilize Google's Gemini~\cite{gemini} to generate the text descriptions. Gemini~\cite{gemini} is an easy-to-access and accurate LLM that can be utilized as an image-to-text model to describe the ground images. We used the Google Gemini API\footnote{\scriptsize{\url{https://ai.google.dev/tutorials/python_quickstart}}}, with two inputs: the ground image and a custom-designed prompt. 
\supp{For more details of the prompts, please refer to the supplementary material.} A randomly sampled image-text pair is shown in~\cref{fig:dataset_samples}.

\begin{table}
\vspace{-10pt}
    \centering
    \footnotesize
    \setlength{\tabcolsep}{1.8 mm}
    \begin{tabular}{c|cc} \toprule
     & VIGOR~\cite{Vigor} & VIGOR\+ (ours) \\ \midrule
    Ground Images & 105,214 & 105,214 \\
    Aerial Images & 90,618 & 105,214 \\
    Layout Maps & N/A & 105,214 \\
    Geographically Splits & $\times$ & $\checkmark$ \\ 
    Text Description of Ground Image & $\times$ & $\checkmark$ \\
    Words per Description & N/A & 49.82 \\
    \bottomrule
    \end{tabular}
    \caption{Statistics comparison between the original VIGOR~\cite{Vigor} datasets and our proposed VIGOR\+.}
    \label{tab:dataset_statistics}
    \vspace{-15pt}
\end{table}

\noindent\textbf{Geographical Dataset Splits: }
One challenge in applying the original VIGOR on the G2A task is data leakage. This leakage is caused by the overlap between the training and testing data, i.e., samples from both sets were captured nearby, often along shared streets. 
To tackle this issue, we adopt a train-test split based on the geographic location of an image within the city. Specifically, we divide each city into northern and southern regions. The north, covering 80\% of the city, is designated for training, while the remaining 20\% in the south is allocated for testing. \cref{fig:dataset_samples} visualizes the new training and testing splits in New York City. Moreover, following the original VIGOR dataset, we also established \textbf{same-area} (training on 4 cities and testing on 4 cities) and \textbf{cross-area} (training on Seattle and New York, testing on San Francisco and Chicago) protocols for comprehensive evaluation purposes. A comparison between the original VIGOR~\cite{Vigor} and our VIGOR\+ is summarized in~\cref{tab:dataset_statistics}. \supp{For more details regarding our VIGOR\+, please refer to our supplementary material.}

\begin{figure*}[!ht]
    \centering
    \vspace{-20pt}
    \includegraphics[page=2,width=1.\textwidth, trim={8.5cm 16.5cm 6.5cm 10.5cm}, clip]{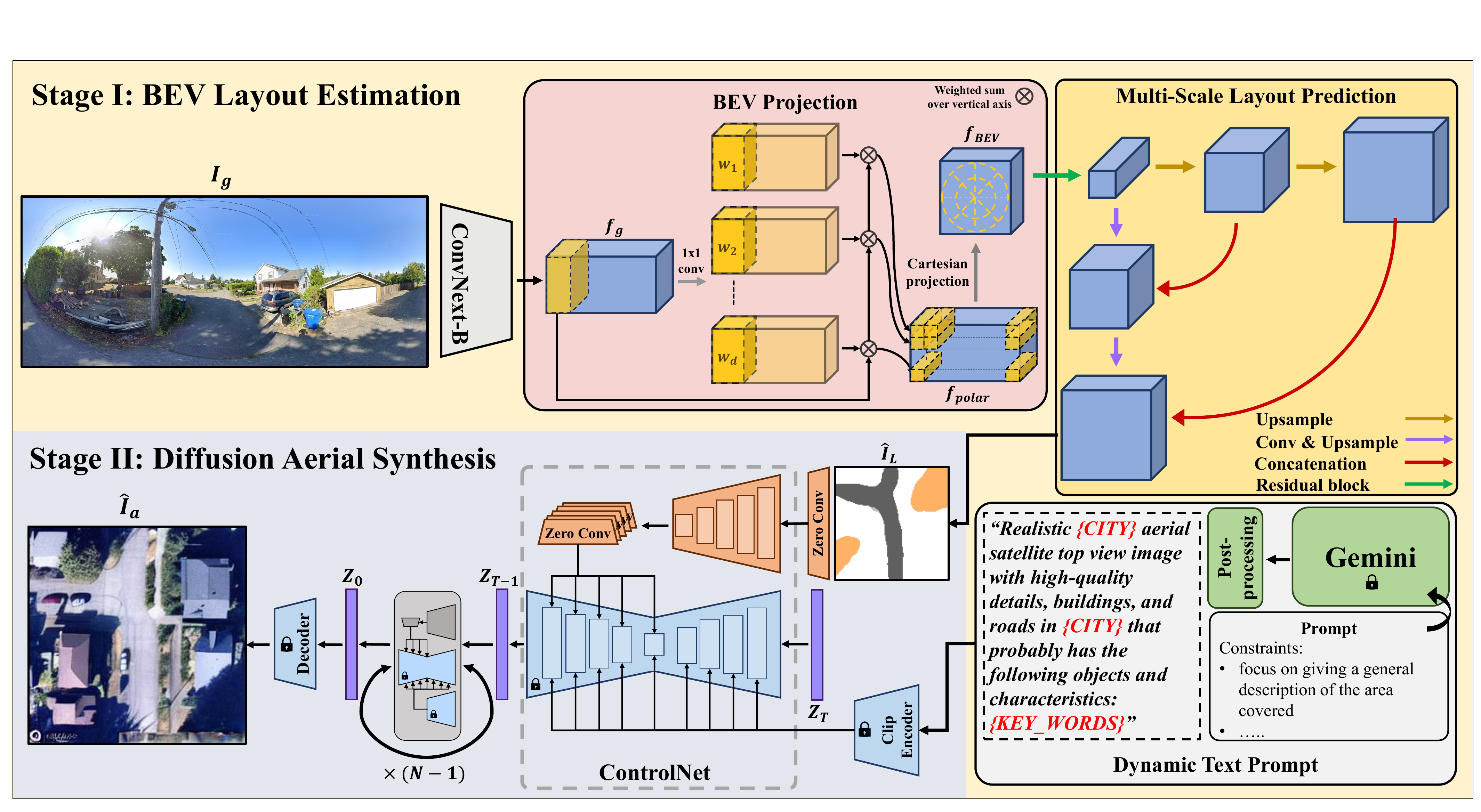}
    \caption{The main architecture of our \ourmodel{}. The first stage is composed of BEV projection and multi-scale layout prediction. Each column in $f_g$ is projected into a polar ray in $f_{BEV}$. The multi-scale network generates the BEV layout map. Then, the second stage synthesizes the aerial image using both $\hat{I}_L$ and the dynamic text prompt. All blocks with a lock symbol indicate a frozen model}
    \label{fig:model}
    \vspace{-15pt}
\end{figure*}

\section{Methodology}
Considering a center-aligned ground-aerial image pair $I_g$ and $I_a$, \ourmodel{} learns the transformation from the ground view to the aerial view through generating an aerial image $\hat{I}_a$ from $I_g$. Directly learning this transformation while preserving visual and geometrical information is challenging, primarily due to the significant change in viewing angle between ground and aerial perspectives. 
To address this challenge, we hypothesize that conditioning geometric priors as an intermediate step improves the synthesis process. Thus, we propose a two-stage model that synthesizes $\hat{I}_a$ by explicitly learning the geometry of $I_a$ from estimating a BEV layout map $\hat{I}_L$ from $I_g$.
Solely depending on the spatial geometric cues from $\hat{I}_L$ would miss textural details. 
To address this, we incorporate text descriptions of the ground image to complement $\hat{I}_L$. These text descriptions are rich in conditioning information that adds realism and fidelity to the generated aerial image.

Our \ourmodel{} model can be formalized as follows,
\begin{equation}
    \hat{I}_a = f\left( h_\phi(I_g),\ \tau(I_g) \right)
    \label{eq:GPG2A}
\end{equation}
In \cref{eq:GPG2A}, the first stage (BEV Layout Estimation) $h$ is parameterized by $\phi$, in which a BEV layout map 
is estimated from the given ground image $I_g$. This layout is expected to share the geometry of $I_a$. $\tau$ is a text extraction module that generates the text description of $I_g$. 
The second stage (Diffusion Aerial Synthesis), $f$, is a generative diffusion model where we condition the estimated BEV layout in addition to the extracted text description from $I_g$. 

\subsection{Stage I: BEV Layout Estimation}
The first stage of \ourmodel{} is responsible for estimating the BEV layout map $\hat{I}_L$ from the input ground image $I_g$. Initially, the ground image undergoes processing through a backbone network, which extracts a latent representation denoted as $f_g \in \mathbb{R}^{c\times h\times w}$, where $c$, $h$, and $w$ are the channel, height, and width dimensions, respectively. For this work, we adopt ConvNeXt-B~\cite{convnext} as our backbone network. Subsequently, we derive a BEV feature map by projecting $f_g$ into the polar space. This BEV feature gets decoded to produce the segmentation layout map $\hat{I}_L$.

Polar transformations have recently found success in both geo-localization \cite{SAFA} and BEV estimation \cite{BEV6,bev7,orienternet}. Therefore, we aim to transform $f_g$ into the polar feature representation $f_{polar}\in\mathbb{R}^{c\times d\times w}$, where $d$ is the introduced depth dimension. $f_{polar}$ maps each column in $f_{g}$ into a polar ray of $d$ cells. Each cell is a result of a dynamic weighted average of its corresponding column in $f_g$. These dynamic weights are introduced by expanding $f_g$ along the depth dimension using 1$\times$1 convolutions, followed by softmax normalization. Thus, as each column in $f_g$ is dynamically weighted $d$ times to produce $d$ cells in the polar ray, we establish a dynamic learnable depth-aware representation of $f_g$, denoted as $f_{polar}$, as visualized in Stage I in ~\cref{fig:model}.

To formalize the dynamic polar projection, we define the dynamic weights as $W_{depth} = g_\theta(I_g) \in \mathbb{R}^{c\times d\times h\times w}$, where $g$ represents the 1$\times$1 convolution network parameterized by $\theta$, which expands $f_g$ along the new depth dimension. To compute the weighted average for all columns in $f_g$, we compute the element-wise multiplication of $f_g$ and $W_{depth}$ for each $d$ in the depth dimension, by splitting $W_{depth}$ into $d$ matrices of shape $[c\times h\times w]$. Subsequently, we sum over the $h$ dimension and concatenate all $d$ multiplication results to obtain $f_{polar}$ with shape $[c\times d \times w]$. The extraction of $f_{BEV}$ can be formulated in ~\cref{eq:BEV_proj}, where $\sigma$ is softmax normalization along the $h$ dimension, and the operation is done for all elements $d_i$ in the $d$ dimension. Moreover, $f_{polar}$ is resampled into a Cartesian grid to derive $f_{BEV} \in \mathbb{R}^{c\times k\times k}$, for any arbitrary choice of $k\in\mathbb{Z}^+$,
\begin{equation}
    f_{polar} = \sum_{h} (f_g \cdot \sigma(W_{depth})) \quad \forall d_i \in d.
  \label{eq:BEV_proj}
\end{equation}

To decode $f_{BEV}$ into a segmentation map $\hat{I}_L$ which will be used in the second stage, we propose the multi-scale layout prediction module, as illustrated in ~\cref{fig:model}. The decoding network is composed of residual blocks ~\cite{resnet} followed by a multi-scale feature concatenation structure. This design simultaneously refines and upsamples the processed BEV feature map by learning both low- and high-level semantic information.
To train the first stage of \ourmodel{}, we adopt the Dice loss ~\cite{dice} defined as, $L_{Dice} = 1 - \frac{2 \ |\hat{I}_L \cap I_L|}{|\hat{I}_L| + |I_L|}$. 

\subsection{Stage II: Diffusion Aerial Synthesis}
\label{sec:stage2}

\subsubsection{Dynamic Text Prompts}
We leverage the versatility of the diffusion model by incorporating an additional modality, specifically text conditions. These encapsulate the environmental and scenic context of the captured area, improving the synthesized aerial images with elements beyond geometry. However, the raw Gemini descriptions contain minor errors and hallucinations which eventually degrade the quality of the generated aerial images (see~\cref{sec:ablation}).

We employ a text extraction post-processing that filters the text and extracts keywords of interest. The extracted keywords, as well as the prior knowledge, e.g., city name, are combined in a template (see~\cref{fig:model} ``Dynamic Text Prompt'' panel for details). 
We focus only on important details in the raw description by constraining it in the template, naming this process, the "dynamic" text prompt. 
To perform keyword extraction, we adopt a BERT-based off-the-shelf model\footnote{\scriptsize{\url{https://maartengr.github.io/KeyBERT/}}} which utilizes BERT embedding and cosine similarity to identify $m$ $N$-gram phrases that closely resemble the raw text. 
The key phrases are ranked by the Maximal Marginal Relevance (MMR) technique~\cite{MMR} based on their relevance to the text. \supp{Refer to our supplementary material for more information about MMR.}

\subsubsection{Model Training}


In \ourmodel{}'s second stage, the following simplified objective from \cite{ddpm} is used to train the diffusion model.
\begin{equation}
\label{eq:loss_ldm}
L_{LDM} \coloneqq \mathbb{E}_{\textbf{Z}, \epsilon\sim \mathcal{N}(0,1), t}\Bigl[\lVert \epsilon - \epsilon_\theta(t, z_t, \tau(I_g), \hat{I}_L) \rVert^2_2\Bigr],
\end{equation}
where $\textbf{Z}$ is the latent representation of the images generated from the pre-trained Variational Autoencoder (VAE) from \cite{LDM}. $\epsilon_\theta$ is parameterized by $\theta$ and defined as the time-conditioned U-net~\cite{unet} with our additions, i.e., text extraction module $\tau$ and the BEV layout map $\hat{I}_L$. t is the time step value in the diffusion process.

\section{Experiments}
\label{sec:experiments}

\subsection{Evaluation metrics}
\label{sec:eval}
\begin{figure}[!ht]
\vspace{-10pt}
\centering
    \centering
    \includegraphics[page=3,width=0.47\textwidth, trim={10.5cm 39.5cm 31cm 15cm}, clip]{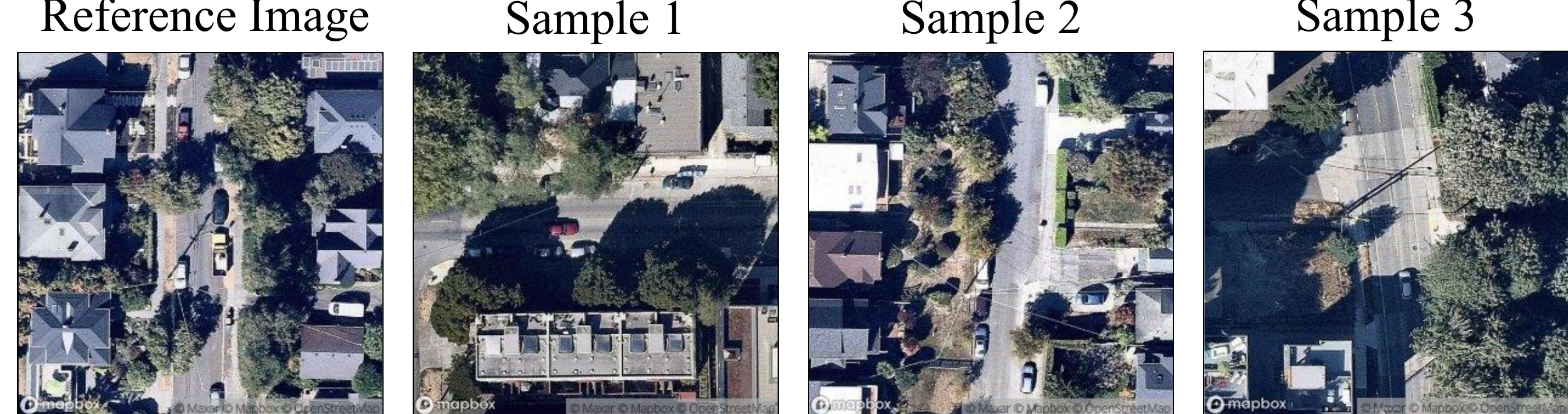}
    \caption{One reference image and three samples for evaluation metrics comparisons.}
    \label{fig:similarity_illustration}
    \vspace{-15pt}
\end{figure}

\begin{table}[!ht]
    \vspace{-3pt}
    \centering
    \setlength{\tabcolsep}{5 mm}
    \footnotesize
    \begin{tabular}{c|ccc} \toprule
        Metrics & Sample 1 & Sample 2 & Sample 3 \\ \midrule
        PSNR $\uparrow$ & 8.587 & 8.302 & 8.554 \\
        SSIM $\downarrow$ & 0.062 & 0.052 & 0.060 \\
        LPIPS $\downarrow$ & 0.753 & 0.722 & 0.784 \\
        $Sim_s$ $\downarrow$ & 0.510 & 0.362 & 0.409 \\
        $Sim_c$ $\downarrow$ & 0.467 & 0.370 & 0.416 \\ \bottomrule
    \end{tabular}
    \caption{Evaluation metrics comparison between the sample images and the reference in~\cref{fig:similarity_illustration}. Existing methods (PSNR, SSIM, LPIPS) can hardly capture the similarity in aerial images. $\uparrow$ means higher better. $\downarrow$ means lower better.}
    \label{tab:eval_metrics}
    \vspace{-10pt}
\end{table}

In the G2A task, popular image quality evaluation metrics such as PSNR, SSIM~\cite{SSIM}, and LPIPS~\cite{LPIPS} are insufficient to evaluate the similarity in aerial images. For illustration, we select one reference image and three test images as shown in~\cref{fig:similarity_illustration}. Sample 1 shares a different layout (horizontal street) than samples 2 and 3 (vertical street). We measure the similarity between each sample and the reference image using the aforementioned metrics in ~\cref{tab:eval_metrics}. PSNR, SSIM, and LPIPS do not reflect the similarities as all three metrics show minor differences. This is because these metrics either only estimate pixel-level similarity (PSNR and SSIM) or lack knowledge of aerial image data (LPIPS). 

To address the above-mentioned issue, we propose a new approach to evaluate our proposed methods by using one of the state-of-the-art cross-view geo-localization (CVGL) model~\cite{SAFA} to estimate the similarity between real and synthesized aerial images. The goal of CVGL is to minimize the distance between matched aerial-ground pairs and maximize the distance between the unmatched ones. Formally, denote $f^a$, $f^g$, and $\hat{f}^a$ as the $L_2$ normalized features for real aerial images, corresponding ground images, and synthesized aerial images, respectively from a well-trained CVGL model (i.e. SAFA~\cite{SAFA}). If the synthesized aerial image is realistic and geometrically preserved, the distance between $f^a$ and $\hat{f}^a$ should be small and we name it same-view similarity metric  ($Sim_s$) which are formally defined as follows,

\begin{equation}
    Sim_{s} = \frac{1}{N} \sum_{i=1}^{N} \frac{2-2 \times (f^{a} \cdot \hat{f}^a)}{4},
    \label{eq::similarity_eq}
\end{equation}

\noindent where $N$ is the number of samples. Correspondingly, we also evaluate the similarity between $f^g$ and $\hat{f}^a$ and we name it cross-view similarity metric ($Sim_c$) which can be easily obtained by replacing $f^a$ into $f^g$ in~\Cref{eq::similarity_eq}. To extract the features, we train the SAFA~\cite{SAFA} on the training set of VIGOR\+. In~\cref{tab:eval_metrics}, $Sim_s$ and $Sim_c$ shows that sample 2 and sample 3 are closer to the reference image than sample 1. This indicates its efficacy in evaluating the synthesized images in this task. Besides $Sim_s$ and $Sim_c$, we also adopt a customized FID~\cite{FID} score, namely $\text{FID}_\text{SAFA}$ that leverages the features ($f^a$ and $\hat{f}^a$) to evaluate the divergence between real images and synthesized images. \supp{For more details, please refer to the supplementary material.}


\subsection{Quantitative Results}

\begin{table}[!ht]
    \centering
    \footnotesize
    \footnotesize
    \setlength{\tabcolsep}{0.4 mm}
    \begin{tabular}{c|cccccc}
    \toprule 
    \multirow{2}{*}{Method} 
    & \multicolumn{3}{c}{Same-area} & \multicolumn{3}{c}{Cross-area} \\
    \cmidrule(lr){2-4}
    \cmidrule(lr){5-7}
    & $Sim_s \downarrow$ & $Sim_c \downarrow$ & $\text{FID}_\text{SAFA} \downarrow$ & $Sim_s \downarrow$ & $Sim_c \downarrow$ & $\text{FID}_\text{SAFA} \downarrow$ \\ \midrule
    X-seq & 0.392 & 0.438 & 0.411 & 0.392 & 0.454 & 0.570 \\
    X-fork & 0.341 & 0.423 & 0.151 & 0.372 & 0.445 & 0.357 \\
    $\text{ControlNet}^{\dag}$ & 0.435 & 0.415 & 0.154 & 0.446 & 0.405 & 0.386 \\
    $\text{ControlNet}^{\ddag}$ & 0.369 & 0.412 & 0.110 & 0.409 & 0.420 & 0.220 \\
    \rowcolor{lightgray} \ourmodel{} (ours) & 0.295 & 0.402 & 0.079 & 0.333 & 0.392 & 0.197 \\
    \bottomrule
    \end{tabular}
    \caption{Same-area and cross-area benchmark results between our proposed \ourmodel{} with baseline methods on our VIGOR\+. The best results are highlighted in a gray background. $\dag$ indicates that a fixed text prompt is used for training the ControlNet. $\ddag$ indicates training the ControlNet with the dynamic text prompts proposed in this paper. $\downarrow$ indicates that the lower value is better.}
    \label{tab:main_result}
    \vspace{-10pt}
\end{table}

To evaluate our \ourmodel{}, we benchmark it on the proposed VIGOR\+ dataset in both same-area and cross-area protocols. As discussed in~\cref{sec:eval}, we rely on the $Sim_c$, $Sim_s$, and $FID$ score for comparison. We choose ControlNet~\cite{controlnet}, X-fork, and X-seq~\cite{regmi2018cross} as the baseline methods. For a fair comparison, two versions of ControlNet were evaluated, one with a constant prompt condition and another with our proposed dynamic prompt. To our best knowledge, X-fork and X-seq are the only models to tackle the G2A task. The experimental results are presented in~\cref{tab:main_result} in which the left panel shows the same-area results and the right panel shows the cross-area results. Our proposed \ourmodel{} achieves the best results among all the baseline methods in both same-area and cross-area experiments. Notably, the $Sim_s$ and $\text{FID}_\text{SAFA}$ of our \ourmodel{} are substantially better than other baseline methods. On the other hand, ControlNet~\cite{controlnet} does not outperform the GAN-based X-fork~\cite{regmi2018cross} in $Sim_s$ and $Sim_c$. This illustrates that without the input of the geometric prior, i.e., BEV layout maps, ControlNet can hardly infer the ground to aerial view changes. This observation supports our two-stage pipeline which divides the BEV estimation and aerial synthesis. A clear improvement in ControlNet can be noticed when using the dynamic text prompt, which validates the use of our text extraction module. In the cross-area experiment, we notice that X-seq has a larger $Sim_c$ and $\text{FID}_\text{SAFA}$ score. This might be attributed to the overfitting issue in this GAN-based method that cannot generalize to unseen data. However, our proposed \ourmodel{} can still maintain an outstanding performance in the cross-area experiments. \supp{For conventional PSNR, SSIM, and LPIPS scores, please refer to our supplementary materials.}

\subsection{Qualitative Results}
\label{sec::qualitative_exp}
\begin{figure}[!ht]
\vspace{-15pt}
    \centering
    \includegraphics[page=4,width=0.48\textwidth, trim={0.5cm 20.2cm 25cm 2cm}, clip]{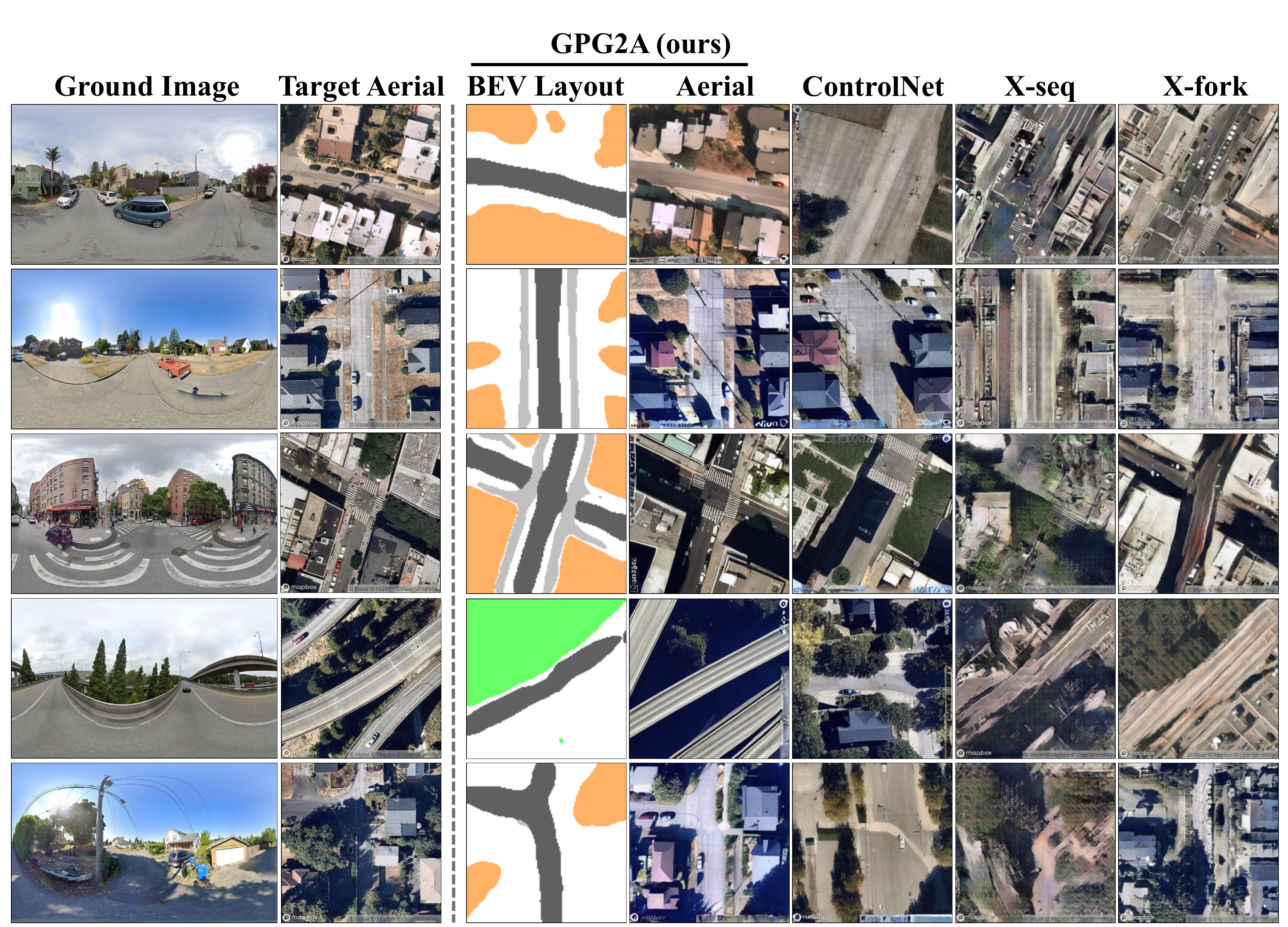}
    \caption{Same-area qualitative comparison. From left to right are ground images, target aerial images, ours synthesized BEV layouts and aerial images, ControlNet~\cite{controlnet}, X-seq~\cite{regmi2018cross}, and X-fork~\cite{regmi2018cross}.}
    \label{fig:same_view_images}
    \vspace{-5pt}
\end{figure}

\noindent\textbf{Same-Area Experiment:} Some randomly selected samples are visualized in~\cref{fig:same_view_images}. For our \ourmodel{}, we present both generated aerial images and predicted BEV layout maps. Notably, the synthesized aerial images and BEV layout maps share geometric structures, providing empirical support to our hypothesis that the BEV map prior would lead to better synthesis. Compared to other baselines, especially in the first and the fourth example in~\cref{fig:same_view_images}, \ourmodel{} preserves geometry and generates high-quality aerial images with details. However, ControlNet has some details, e.g., roads and trees, but lacks geometric correspondence. X-fork and X-seq generate blurry images without details.

\begin{figure}[!ht]
\vspace{-5pt}
    \centering
    \includegraphics[page=5,width=0.48\textwidth, trim={0.5cm 20.2cm 25cm 2cm}, clip]{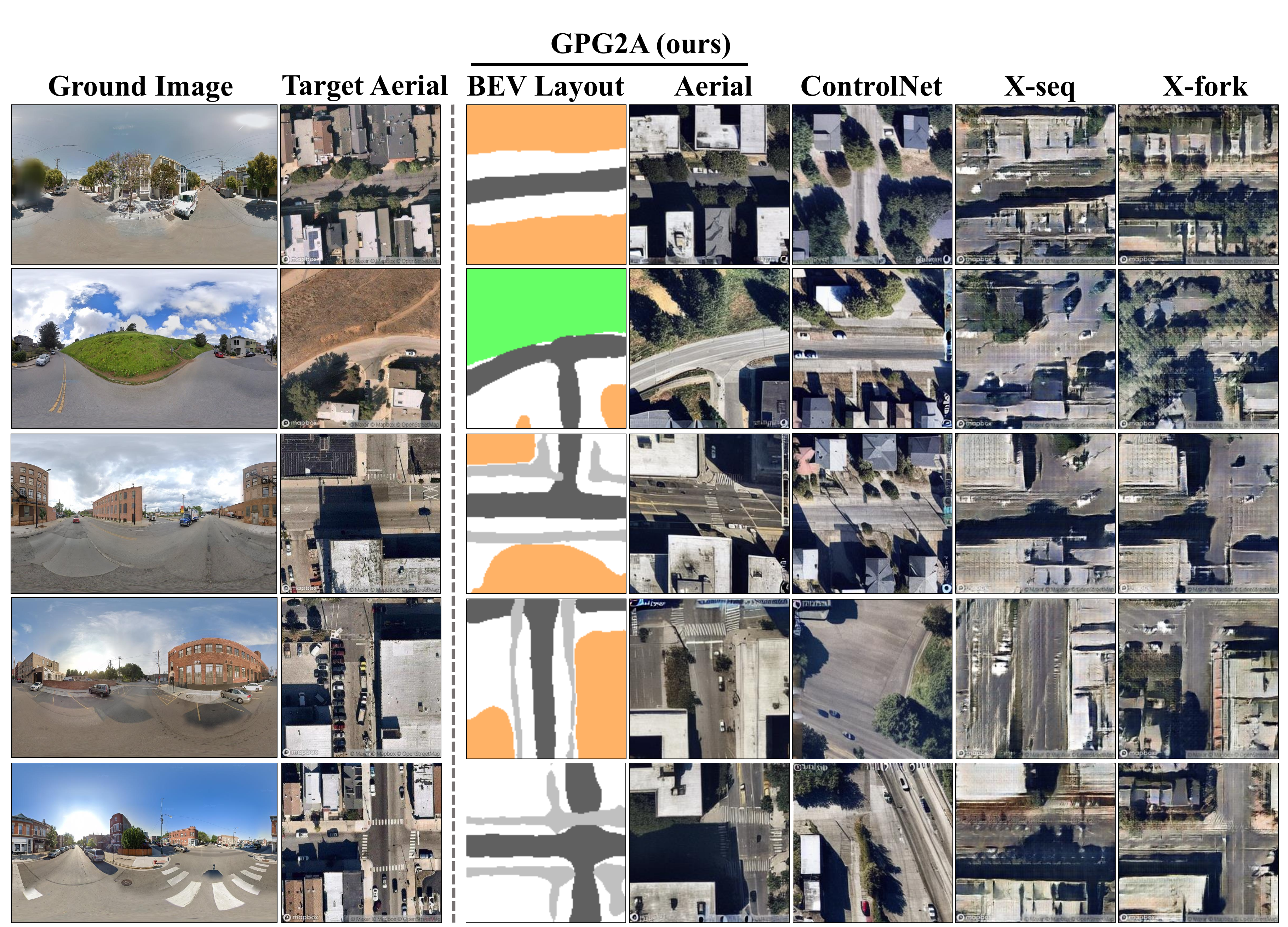}
    \caption{Cross-area qualitative comparison. From left to right are ground images, target aerial images, ours synthesized BEV layouts and aerial images, ControlNet~\cite{controlnet}, X-seq~\cite{regmi2018cross}, and X-fork~\cite{regmi2018cross}.}
    \label{fig:cross_view_images}
    \vspace{-15pt}
\end{figure}

\noindent\textbf{Cross-Area Experiment:} To further validate the generalization of \ourmodel{} on unseen data, we devise a cross-area experiment as visualized in~\cref{fig:cross_view_images}. It is clear to see that the accurate estimation of the BEV layout maps preserves geometric consistency even in unseen scenarios. To be noticed, some disparities appear in environmental details, such as the appearance of buildings (the fifth example in~\cref{fig:cross_view_images}). On the other hand, all other baseline methods generated samples lacking both geometry and details compared with our \ourmodel{}. \supp{For more, qualitative examples and failure cases, please refer to our supplementary materials.}

\subsection{Ablation Studies}
\label{sec:ablation}

\begin{table}[!ht]
\vspace{-10pt}
    \centering
    \footnotesize
    \setlength{\tabcolsep}{0.65 mm}
    \begin{tabular}{c|cccccc}
    \toprule 
    \multirow{2}{*}{Prompt}& \multicolumn{3}{c}{Same-area} & \multicolumn{3}{c}{Cross-area} \\
    \cmidrule(lr){2-4}
    \cmidrule(lr){5-7}
    & $Sim_s \downarrow$ & $Sim_c \downarrow$ & $\text{FID}_\text{SAFA} \downarrow$ & $Sim_s \downarrow$ & $Sim_c \downarrow$ & $\text{FID}_\text{SAFA} \downarrow$ \\ \midrule
    Raw & 0.383 & 0.425 & 0.123 & 0.384 & 0.412 & 0.227 \\
    Constant & 0.323 & 0.418 & 0.131 & 0.362 & 0.407 & 0.259 \\
    City-only & 0.316 & 0.419 & 0.087 & 0.356 & 0.424 & 0.208 \\
    \rowcolor{lightgray} Dynamic & 0.295 & 0.402 & 0.079 & 0.333 & 0.392	& 0.197 \\
    \bottomrule
    \end{tabular}
    \caption{Ablation study of the text prompt in the proposed \ourmodel{}. `Constant' indicates fixing the text prompt. `Raw' stands for using raw text descriptions from Gemini without keyword selection. `City-only' means varying the city name in the prompt. `Dynamic' stands for the proposed dynamic text prompt.}
    \label{tab:text_ablation}
    \vspace{-10pt}
\end{table}

\noindent\textbf{Text prompt: } Text prompts provide important contextual details for \ourmodel{}, as mentioned in~\cref{sec:stage2}. In this experiment, we ablate different types of prompts in the training phase to demonstrate the effectiveness of our dynamic prompt design. We study three additional types of prompts: the constant prompt, a fixed generic text prompt; the raw prompt, which directly applies the Gemini output; and the city-only prompt, which only varies the city name. The experiment results are presented in~\cref{tab:text_ablation}. First, the "Raw" prompt achieves the worst results. We attribute this to the lengthy raw text from Gemini (potentially with hallucination), resulting in a noisy signal to the model. 
It is noted that the "constant" prompt is better than the "Raw" prompt in $Sim_s$ and $Sim_c$ but worse in $\text{FID}_\text{SAFA}$. This degradation might be attributed to the absence of surrounding information from ground images. This also reveals the importance of our dynamic prompt which boosts the model in all evaluation metrics on both same-area and cross-area settings.

\begin{table}
    \centering
    \footnotesize
    \setlength{\tabcolsep}{5 mm}
    \begin{tabular}{c|ccc} \toprule
    Method & $Sim_s \downarrow$ & $Sim_c \downarrow$ & $\text{FID}_\text{SAFA} \downarrow$ \\ \midrule
         $f_{bev}$ & 0.465 & 0.478 & 0.426 \\
        Ours & 0.295 & 0.402 & 0.079 \\ \bottomrule
    \end{tabular}
    \caption{Ablation study on the stage II input. `$f_{bev}$' indicates directly applying the $f_{bev}$ to stage II.}
    \label{tab:fbev_ablate}
    \vspace{-10pt}
\end{table}

\noindent\textbf{BEV layout input: }
To verify our choice of using the BEV layout map as conditioning, we compare it to the BEV feature $f_{BEV}$ on the same-area experiment. As indicated in~\cref{tab:fbev_ablate}, directly applying $f_{bev}$ to stage II would significantly degrade the performance in all metrics which is a firm support to our assumption, and demonstrates the effectiveness of the intermediate BEV layout extracted by the multi-scale layout prediction network.

\begin{table}
    \footnotesize
    \setlength{\tabcolsep}{2.5 mm}
    \centering
    \begin{tabular}{c|ccccc} \toprule
         FOV & \multicolumn{2}{c}{BEV Accuracy} & \multicolumn{3}{c}{Synthesis Quality} \\ 
         \cmidrule(lr){1-1}
         \cmidrule(lr){2-3}
         \cmidrule(lr){4-6}
    & Avg F1 & mIoU & $Sim_s \downarrow$ & $Sim_c \downarrow$ & $\text{FID}_\text{SAFA} \downarrow$ \\ \midrule
      $90^{\circ}$  & 0.259 & 0.149 & 0.413 & 0.414 & 0.290  \\
      $180^{\circ}$  & 0.411 & 0.258 & 0.385 & 0.406 & 0.181  \\
      $270^{\circ}$  & 0.458 & 0.297 & 0.369 & 0.404 & 0.143  \\
      $360^{\circ}$  & 0.565 & 0.394 & 0.295 & 0.402 & 0.079  \\ \bottomrule
    \end{tabular}
    \caption{Ablation study on limited FOV input ground images. ``BEV Accuracy'' represents BEV layout prediction accuracy from Stage I and ``Synthesis Quality'' represents the quality of the synthesized aerial image from Stage II.}
    \label{tab:limited_FOV}
    \vspace{-15pt}
\end{table}

\noindent\textbf{Limited FOV ground images: }In previous experiments, we assume the ground images are panoramic. In fact, limited field-of-view (FOV) images are more accessible~\cite{seqgeo}. Thus, we extend stage I of \ourmodel{} to predict BEV layout maps from limited FOV images. As shown in~\cref{tab:limited_FOV}, we use the Average F1 and IoU to evaluate the prediction accuracy in stage I, and $Sim_s$,  $Sim_c$, and $\text{FID}_\text{SAFA}$ to evaluate the synthesis quality in stage II. As shown in~\cref{tab:limited_FOV}, with the increase of FOV, both stages improved. It is noteworthy that by comparing \cref{tab:limited_FOV} and \cref{tab:main_result}, \ourmodel{} is still better than X-seq and X-fork with $180^{\circ}$ FOV input in $Sim_c$ and comparable results on $Sim_s$ and $\text{FID}_\text{SAFA}$. \supp{For qualitative results, please refer to our supplementary materials.}
\section{Applications}

To further evaluate our \ourmodel{} model and validate G2A image synthesis across domains, we apply it to two real-world applications: 1) Data augmentation for geo-localization and 2) Sketch-based region search.


\subsection{Data Augmentation for Geo-localization}

\begin{table}[!ht]
\vspace{-15pt}
    \centering
    \footnotesize
    \setlength{\tabcolsep}{3.3 mm}
    \begin{tabular}{cc|cccc}
    \toprule  
     & $p_\circ$ & R@1 $\uparrow$ & R@5$\uparrow$ & R@10$\uparrow$ & R@1\%$\uparrow$ \\
    \toprule
    \multirow{4}{*}{\rotatebox{90}{same-area}} & 0 & 83.58\%         & 93.72\%          & 96.08\%          & 99.27\%  \\
      & 0.4 & 86.33\%           & 94.76\%           & 96.27\%          & \textbf{99.38}\% \\
    & 0.6 & \textbf{86.84}\%  &\textbf{94.98}\%  & 96.30\%           & 99.34\%  \\
    & 0.8 & 85.77\%           & 94.71\%           & \textbf{96.44}\% & 99.30\%  \\ \midrule
    \multirow{4}{*}{\rotatebox{90}{cross-area}} & 0 & 50.03\%         & 70.17\%          & 77.70\%          & 94.03\%  \\
     & 0.4 & 51.55\%          & 71.96\%          & 78.18\%          & 94.10\%  \\
    & 0.6 & \textbf{52.84}\% & \textbf{72.32}\% & \textbf{78.45}\% & \textbf{94.19}\%  \\
    & 0.8 & 50.11\%          & 70.98\%          & 77.60\%          & 93.99\%  \\
    \bottomrule
    \end{tabular}
    \caption{Results of our data augmentation on SAFA. $p_\circ=0$ indicates no augmentation is applied. It is noticeable that our augmentation can improve in both same-area and cross-area performance.}
    \label{tab:app_samearea}
    \vspace{-15pt}
\end{table}

To develop robust CVGL models, many data augmentation techniques have been proposed~\cite{aug1,wilson2021visual,aug2, geodtr+}. In this application, we propose to apply aerial images generated by GPG2A in the Mixup augmentation method~\cite{mixup} to train robust CVGL models. 
The mixup augmentation can be defined as follows,
\begin{equation}
    \label{eq:mixup}
    \hat{x} =
    \begin{cases} 
       \lambda \ x_{fake} + (1-\lambda) \ x_{real} & p\leq p_\circ \\
        x_{real} & p>p_\circ,
   \end{cases}
\end{equation}
where $\hat{x}, \ x_{fake}, \text{and} \ x_{real}$ are the augmented, generated (\ourmodel{} output), and real aerial images, respectively. $\lambda$ determines the mixup strength. $p_\circ$ is the probability of applying the augmentation.
We apply this augmentation on SAFA~\cite{SAFA} which is a well-known CVGL model. 
The model was trained in New York and Seattle subsets, and evaluated in both same-area and cross-area settings. 

\cref{tab:app_samearea} shows the performance of our data augmentation in the same-area and cross-area settings. We evaluate the performance using recall accuracy at top K (R@K)~\cite{SAFA}, which measures the likelihood that the ground truth aerial image ranks within the top K predictions. Overall, the augmentation improved performance across all metrics. Specifically, on R@1 the data augmentation brings $3.26\%$ and $2.81\%$ improvements to same-area and cross-area tests respectively while $p_\circ=0.6$. We also notice that the performance decreases while $p=0.8$. This might indicate the lack of convergence of the model because of the stronger augmentation. \supp{Please refer to our supplementary material for experiments on more recent CVGL models.}

\subsection{Sketch-based Region Search}

\begin{figure}[!ht]
\vspace{-10pt}
    \centering
    \includegraphics[page=16,width=0.47\textwidth, trim={2.9cm 25.5cm 37.9cm 8.6cm}, clip]{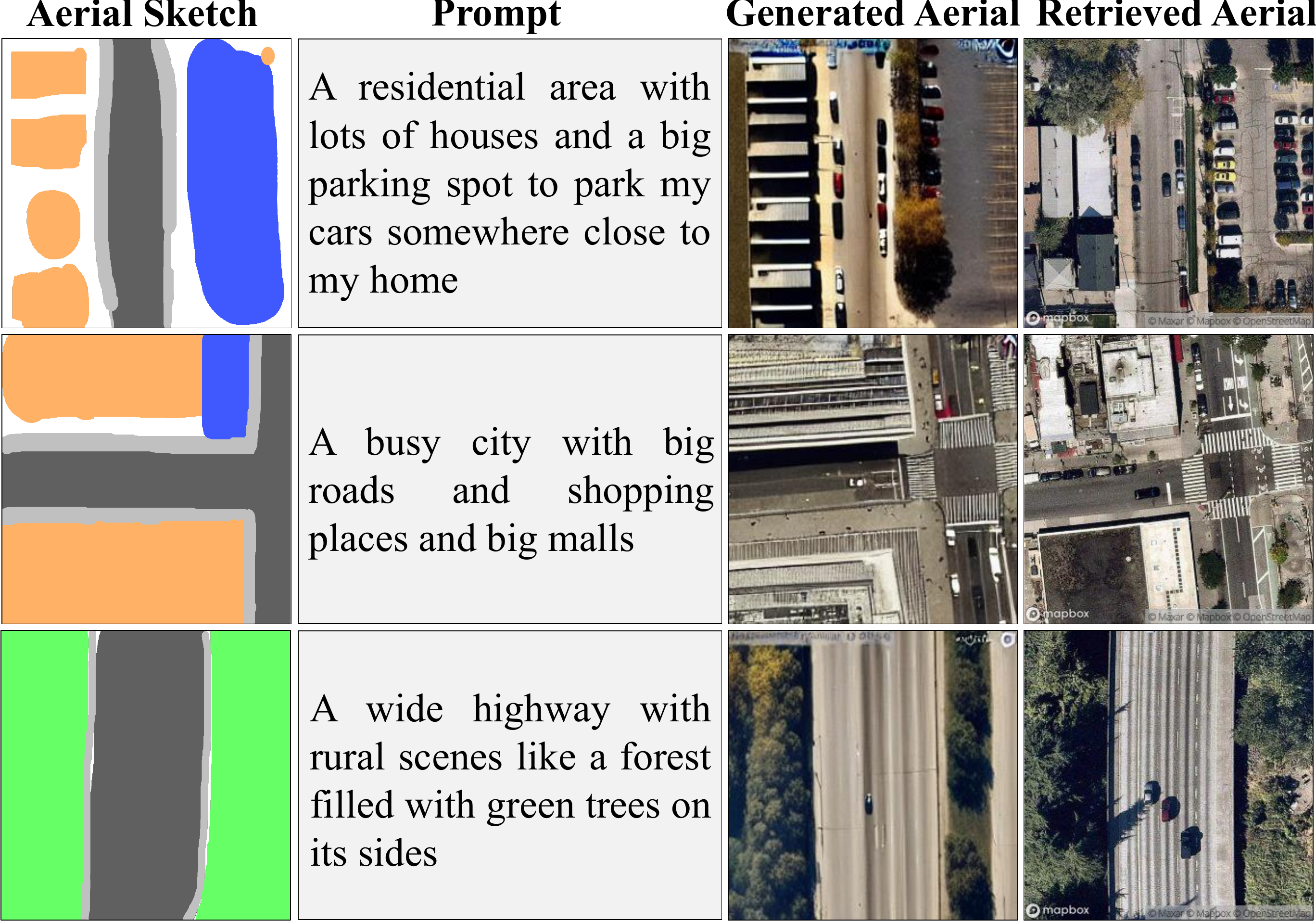}
    \caption{Synthesis and retrieval results of the sketch-based region search application. Each color in the layout sketch represents a class as follows: orange, black, grey, blue, and green reflect buildings, streets, sidewalks, parking lots, and trees, respectively.}
    \label{fig:sketch1}
    \vspace{-10pt}
\end{figure}

Aerial image search is one of the most challenging tasks in remote sensing \cite{sketch2}, particularly when a query image is represented by mental maps, such as hand-drawing sketches~\cite{sketch1} without low-level details. 
In this task, we assume that the user has given a hand-drawn sketch and a text description of the surrounding environment. The goal is to find similar aerial images from a geo-tagged imagery database. In this way, the user can find the point of interest by using only sketches and descriptions. 
Specifically, we use the second stage of \ourmodel{} to synthesize a fake aerial image of the region using the sketch and the text description. Then, we retrieve the most similar aerial image from a reference database by calculating the closest latent features in Euclidean distance. To achieve this, a pre-trained SAFA model is adopted to extract latent features. 

\cref{fig:sketch1} illustrates 3 retrieval results. The analyzed samples were designed to be diverse, with different scenery and objects. For example, a parking lot was added in the first sample, while a highway was included in the third sample. Both the synthesized and retrieved images showed strong correspondence with the aerial sketches and descriptions. To further evaluate this pipeline, we conducted a survey that asked 61 volunteers to identify similarities between 5 groups of the input (aerial sketch and text prompt) and three different aerial images (corresponding top-1 retrieved aerial image, the 5th retrieved aerial image, and a random aerial image). The results show that $66\%$ of the volunteers believe the top-1 retrieved images correspond to the input aerial sketch and text prompt. This number drops to $60\%$ in the 5th retrieved aerial image. While only $24\%$ of the people think random aerial images are similar to the input. It is noteworthy that visualization of aerial images from sketches boosts the search explainability. Most previous works~\cite{sketch1,sketch2} aim to find a common latent space between sketches and aerial images, which lacks interpretability. \supp{For more details, please refer to supplementary material.}

\section{Conclusion and Future Works}
In this paper, we propose \ourmodel{} which is a two-stage model that generates geometry-preserved aerial images from ground images by conditioning on predicted BEV layouts and text descriptions. To alleviate the problem of lacking datasets for benchmarking, we propose VIGOR\+, which is built upon the VIGOR~\cite{Vigor} dataset with newly collected aerial images, BEV layout maps, and text descriptions. Our proposed \ourmodel{} substantially outperforms existing baselines on the VIGOR\+ dataset. Additionally, we apply our \ourmodel{} on two downstream tasks to show its potential practical application. 

As a novel research field, there are many opportunities to advance this research. One of the directions is to investigate fusion from ground videos to synthesize aerial images. Another one is to research more fine-grained conditioning techniques to generate more diverse aerial images, e.g., different seasons and weather conditions.

{\small
\bibliographystyle{ieee_fullname}
\bibliography{main}
}

\clearpage

{\Large \textbf{Supplementary Material}}

\section{Introduction}
In this supplementary material, we provide more information on our implementation details (\cref{sec::implemenataion}). As mentioned in our main script, we presented the evaluation results on traditional SSIM and PSNR evaluation metrics (\cref{sec::SSIM_PSNR}). Furthermore, we visualize the output of stage I of \ourmodel{} in the limited FOV experiment (\cref{sec::limitedFOV}). We present more qualitative samples of our proposed GPG2A in the same-area and cross-area experiments (\cref{sec::qualitative}). Additionally, we analyze some failure synthesis cases (\cref{sec::failure}). More aerial, ground, text, and BEV layout map samples from our VIGOR\+ dataset are presented (\cref{sec::dataset}) along with the description of the text data used in this research (\cref{sec::text}). We also provide additional experiments, results, and more details of the two downstream applications (\cref{secc::app1} and \cref{secc:app2}). Finally, we discuss the limitations (\cref{sec::limitation}) and the societal impact (\cref{sec::social_impact}) of our paper.

\section{Implementation Details}
\label{sec::implemenataion}

We implemented the \ourmodel{} in Pytorch~\cite{pytorch}. In stage I, we trained the model with a batch size of $128$ and Adam~\cite{kingma2014adam} optimizer with the learning rate set to $0.0001$. In the BEV projection step, the depth dimension $d$ was set to 64, and in the Cartesian projection, $k$ was set to 32. Thus, the polar feature $f_{polar}$ was resampled into a Cartesian grid to derive $f_{BEV} \in \mathbb{R}^{c\times k\times k}$ where $c$ is the latent channel dimension and we set it to 256 in our experiments. We use the Torchvision~\footnote{\scriptsize{\url{https://pytorch.org/vision/main/models/convnext.html}}} official implementation of ConvNext-B~\cite{convnext} with ImageNet~\cite{imagenet} pre-trained weights for the backbone feature extractor. The output of stage I is rendered in pixel space as input for stage II. In stage II, we use the Hugging Face Diffuser library's implementation of ControlNet 
from~\footnote{\scriptsize{\url{https://huggingface.co/docs/diffusers/using-diffusers/controlnet}}}. To train the model, we used  Adam~\cite{kingma2014adam} optimizer with a learning rate of $0.0001$ and a batch size of $4$. Stage I is trained on an AMD MI50 GPU and stage II is trained on a Nvidia V100 GPU with $20$ epochs and $30$ epochs, respectively.

\section{SSIM, PSNR, and LPIPS Quantitative Results}
\label{sec::SSIM_PSNR}

As mentioned in the main script, existing evaluation metrics such as PSNR, SSIM~\cite{SSIM}, and LPIPS~\cite{LPIPS} are insufficient to estimate the quality of the synthesized aerial images. Thus, we did not show the results of SSIM, PSNR, and LPIPS scores. In this section, we provide a comprehensive study that compares PSNR, SSIM, and LPIPS. The results are presented in~\cref{tab:convention_results}. As indicated in the table, SSIM, PSNR, and LPIPS show minimal variants in both same-area and cross-area experiments, which reflects our point in Sec. 5.1 of the main script that existing evaluation metrics are insufficient to evaluate the quality of the synthesized aerial images. 

\begin{table*}[!h]
    \centering
    \setlength{\tabcolsep}{2 mm}
    \begin{tabular}{c|cccccc} \toprule
    \multirow{2}{*}{Method} & \multicolumn{3}{c}{Same-area} & \multicolumn{3}{c}{Cross-area} \\
    \cmidrule(lr){2-4}
    \cmidrule(lr){5-7}
         & SSIM $\uparrow$ & PSNR $\uparrow$ & LPIPS $\downarrow$ & SSIM $\uparrow$ & PSNR $\uparrow$ & LPIPS $\downarrow$ \\ \midrule
        X-fork~\cite{regmi2018cross} & 0.11 & 10.66 & 0.58 & 0.10 & 10.34 & 0.60 \\
        X-seq~\cite{regmi2018cross} & 0.10 & 10.30 & 0.61 & 0.09 & 10.15 & 0.62 \\
        ControlNet~\cite{controlnet} & 0.09 & 9.62 & 0.66 & 0.09 & 9.90 & 0.66 \\
        $\text{GPG2A}^{\dag}$ & 0.12 & 10.40 & 0.63 & 0.09 & 9.65 & 0.60 \\
        $\text{GPG2A}^{\ddag}$ & 0.10 & 9.59 & 0.62 & 0.08 & 9.42 & 0.60 \\
        GPG2A (ours) & 0.10 & 10.07 & 0.62 & 0.09 & 10.17 & 0.61 \\
    \bottomrule
    \end{tabular}
    \caption{PSNR, SSIM~\cite{SSIM}, and LPIPS~\cite{LPIPS} results from our proposed GPG2A as well as other baselines on our proposed VIGORv2 datasets in same-area and cross-area experiments protocols. $\uparrow$ stands for the higher value better, $\downarrow$ stands for the lower value better. $\dag$ stands for constant text prompt and $\ddag$ stands for city text prompt.}
    \label{tab:convention_results}
\end{table*}

\section{More details about $\text{FID}_\text{SAFA}$}
\label{sec::fid}

Similar to the original FID score~\cite{FID}, $\text{FID}_\text{SAFA}$ leverages the features ($f^a$ and $\hat{f}^a$) to evaluate the divergence between real and synthesized images. However, to better evaluate the quality of aerial images, we choose to use the features extracted by pre-trained SAFA~\cite{SAFA} which yields better feature quality than the original one, especially for aerial images.$\text{FID}_\text{SAFA}$ can be formally written as follows,
 \begin{equation}
     \text{FID}_\text{SAFA} = \lvert\lvert \mu^a - \hat{\mu}^a \rvert\rvert + Tr(\Sigma^a + \hat{\Sigma}^a - 2(\Sigma^a\hat{\Sigma}^a)^{\frac{1}{2}})
 \end{equation}
 \noindent where $\mathcal{N}(\mu^a, \Sigma^a)$ and $\mathcal{N}(\hat{\mu}^a, \hat{\Sigma}^a)$ are multivariate normal distributions estimate from $f^a$ and $\hat{f}^a$.

\section{Limited FOV Qualitative Results}
\label{sec::limitedFOV}
\begin{figure}[!ht]
    \centering
    \includegraphics[page=6,width=0.47\textwidth, trim={8cm 20cm 45cm 8cm}, clip]{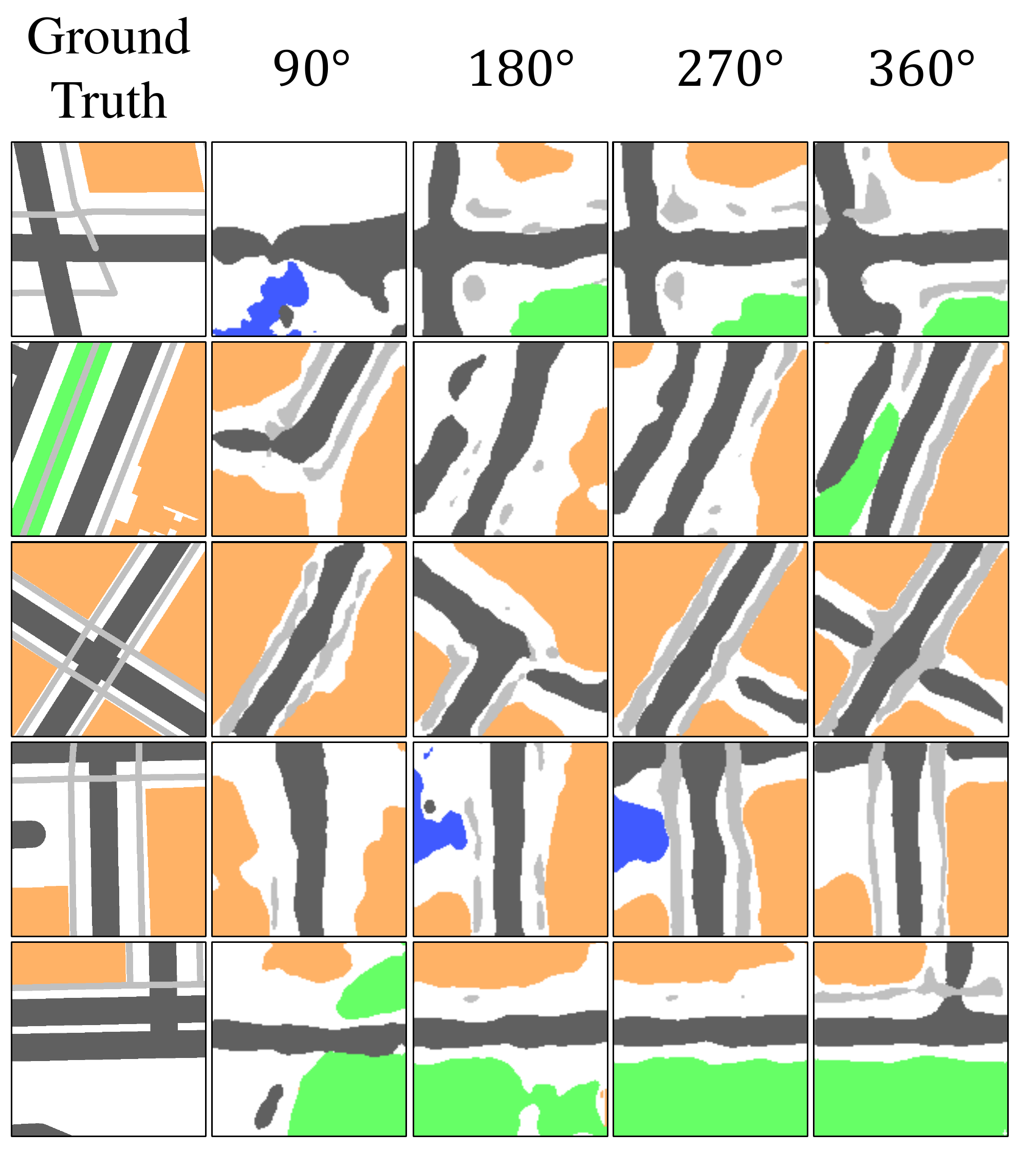}
    \caption{Visualization of the output from stage I of our \ourmodel{} under different limited FOV input ground images. From left to right are ground truth BEV layout map, $90^{\circ}$ FOV, $180^{\circ}$ FOV, $270^{\circ}$ FOV, and $360^{\circ}$ FOV}
    \label{fig:FOV}
\end{figure}

\begin{figure}[!ht]
    \centering
    \includegraphics[page=7,width=0.47\textwidth, trim={13cm 38cm 32cm 16cm}, clip]{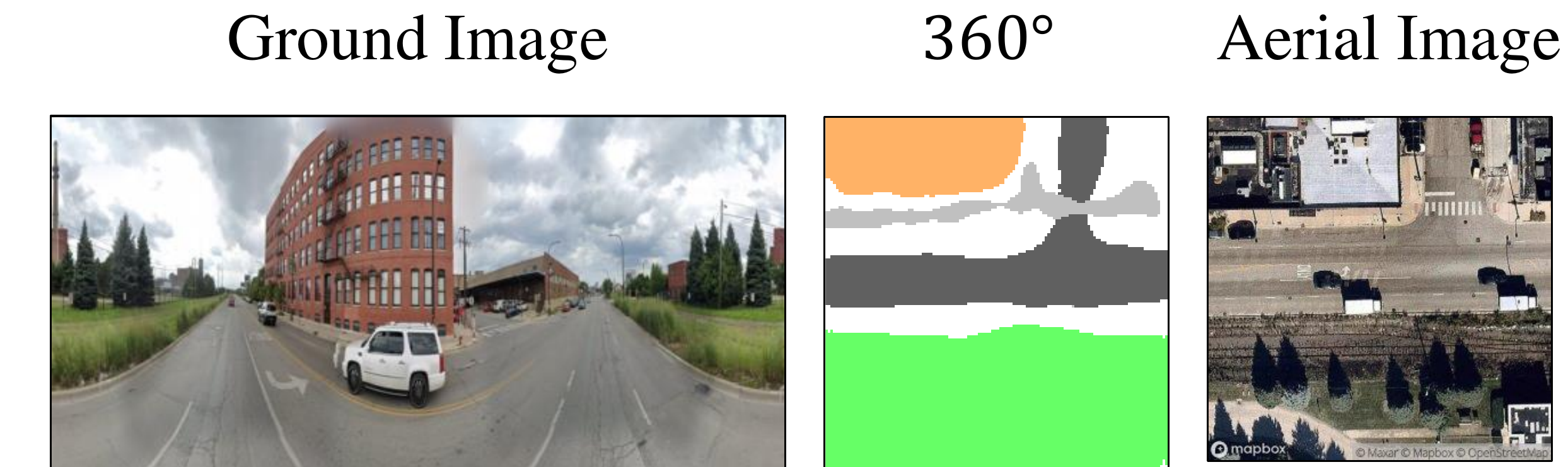}
    \caption{Test sample where stage I inferred trees from the ground image}
    \label{fig:FOV_trees}
\end{figure}

We further extend our analysis of \ourmodel{} in limited FOV ground images by visualizing the predicted BEV layout maps in stage I. As shown in \cref{fig:FOV}, the segmentation results improve as the FOV increases. The samples from the 360$^\circ$ FOV entail more geometric detail of the ground truth. For example, the third sample generated the full intersection while other FOVs did not. Notice how the last sample has a green segmentation class indicating trees, which is not aligned with the ground truth. However, after further investigation, it turned out that there were trees visible in both the ground and aerial counterparts as shown in \cref{fig:FOV_trees}. This indicates that stage I is dynamic and can extrapolate information from the ground image to the BEV layout.  

\section{More Qualitative Results}
\begin{figure*}[!ht]
    \centering
    \includegraphics[page=9,width=0.95\textwidth, trim={0cm 20.2cm 14cm 0cm}, clip]{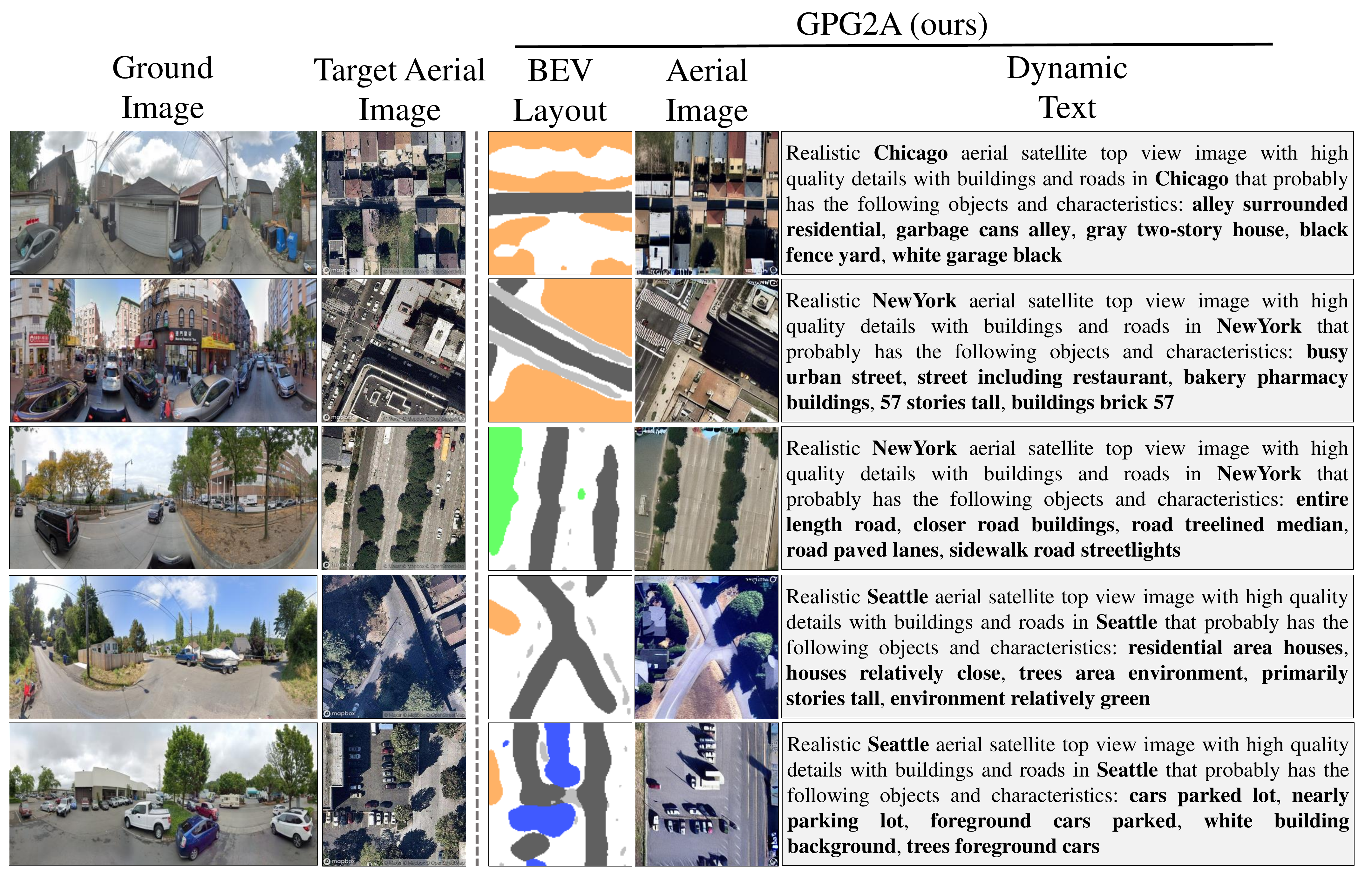}
    \caption{More \textbf{same-area} generated results from our proposed \ourmodel{}. From left to right are the input ground image, target aerial image, predicted BEV layout, synthesized aerial image, and corresponding text prompt.}
    \label{fig:same_area}
\end{figure*}

\begin{figure*}[!ht]
    \centering
    \includegraphics[page=10,width=0.95\textwidth, trim={0cm 20.2cm 14cm 0cm}, clip]{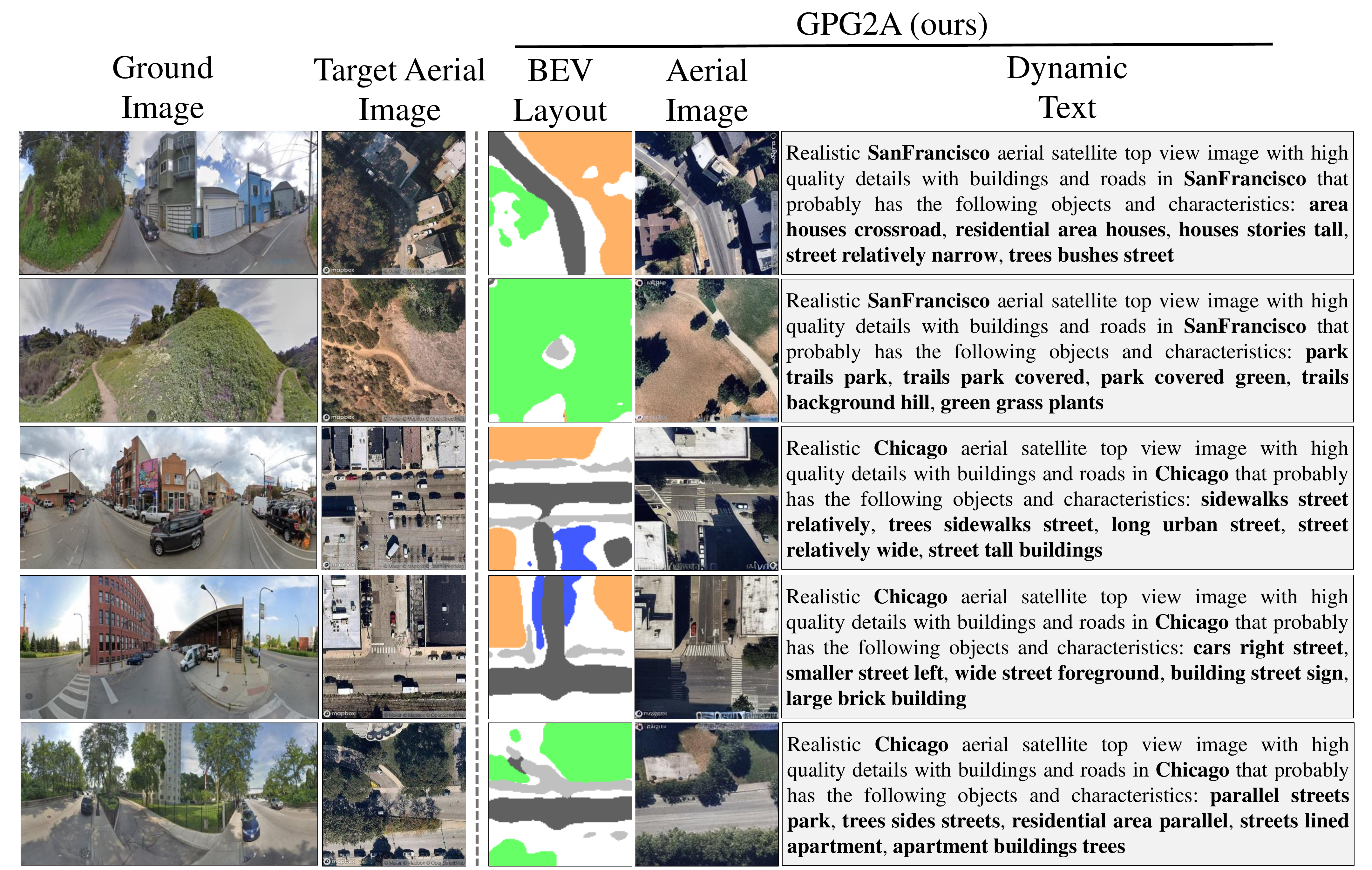}
    \caption{More \textbf{cross-area} generated results from our proposed \ourmodel{}. From left to right are the input ground image, target aerial image, predicted BEV layout, synthesized aerial image, and corresponding text prompt.}
    \label{fig:cross_area}
\end{figure*}

This supplementary material visualizes more diverse synthesis cases of our \ourmodel{} in both the same- and cross-area test cases, as shown in \cref{fig:same_area,fig:cross_area}, respectively. In the same-area results, we can notice how \ourmodel{} generalizes in multiple environments, e.g., residential and urban areas. Also, \ourmodel{} showed great attention to detail, as shown in the tree placement in the third sample, and parking synthesis in the last. Notice how the dynamic text explicitly mentioned these details which guide the synthesis. In the cross-area samples, the geometry was clearly preserved due to the accurate BEV estimates. The dynamic text carried environment details but as the model was trained on different cities, the generated images had some discrepancies with the ground truth. For example, in the first two samples, even though trees were inferred their styles did not match.

\label{sec::qualitative}

\section{Failure cases}
\label{sec::failure}

\begin{figure*}
    \centering
    \includegraphics[page=15,width=1.0\textwidth, trim={1cm 26cm 1cm 6cm}, clip]{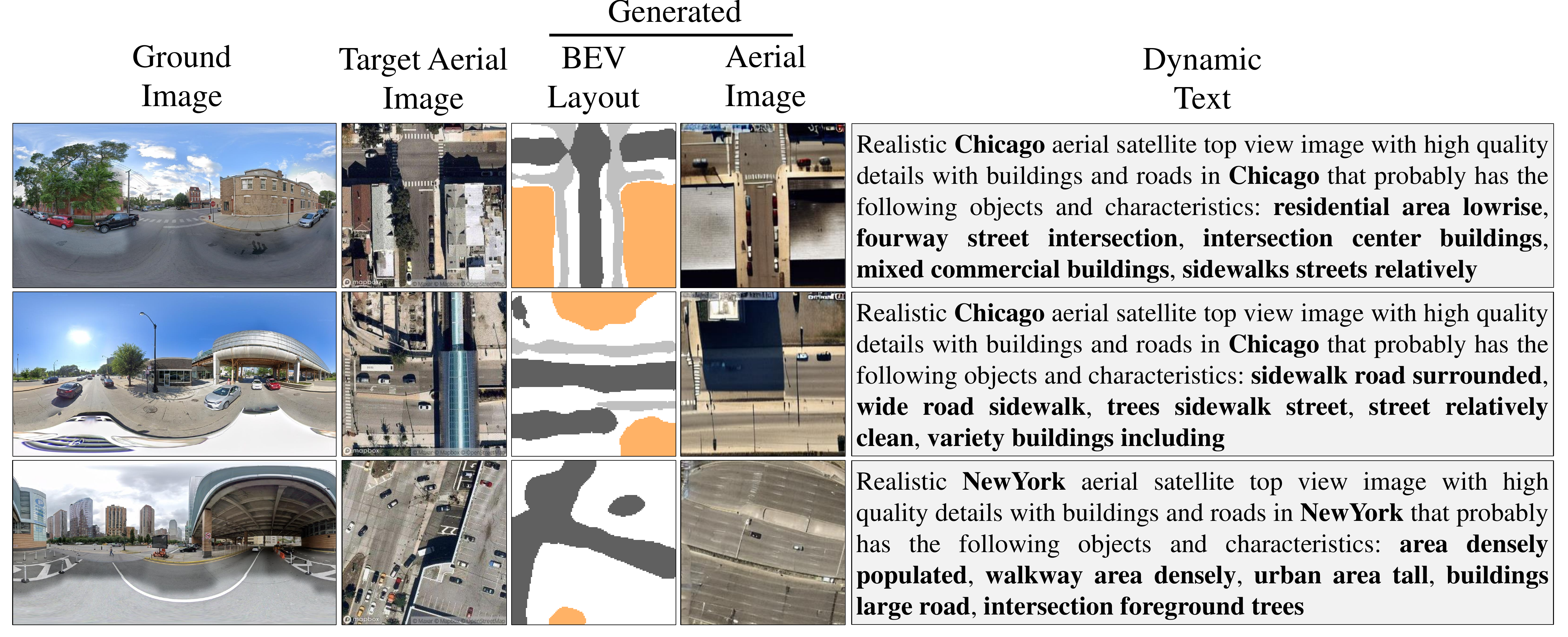}
    \caption{Some randomly sampled failure cases generated by our GPG2A. From left to right are input ground images, target aerial images, predicted BEV layout maps, and generated aerial images.}
    \label{fig:fail}
\end{figure*}

To further study the behavior of our \ourmodel{}, we present some failure cases as shown in~\cref{fig:fail}. The first example (top) keeps the geometric layout as shown in both the predicted BEV layout map and the generated aerial image. However, the generated image lacks details of the buildings and trees. This might be due to the occluded parts in the ground image as a tree was shown to cover the houses. The second (middle) and the third (bottom) examples fail to generate accurate aerial images because of some unseen objects. Specifically, there is a bridge (in the east direction of the ground image) in the ground image which is not captured by stage I of \ourmodel{} as evidenced in the predicted layout image. Except for this bridge, the geometric layout is still preserved. For the last example, there is a two-story parking garage not visible from the ground image which causes the model to fail to capture the geometric layout. Thus, as shown in the generated aerial image, there is no corresponding geometry.

\section{More Dataset Figures}

\begin{figure}[!ht]
    \centering
    \includegraphics[page=8,width=0.45\textwidth, trim={7cm 8cm 45cm 20cm}, clip]{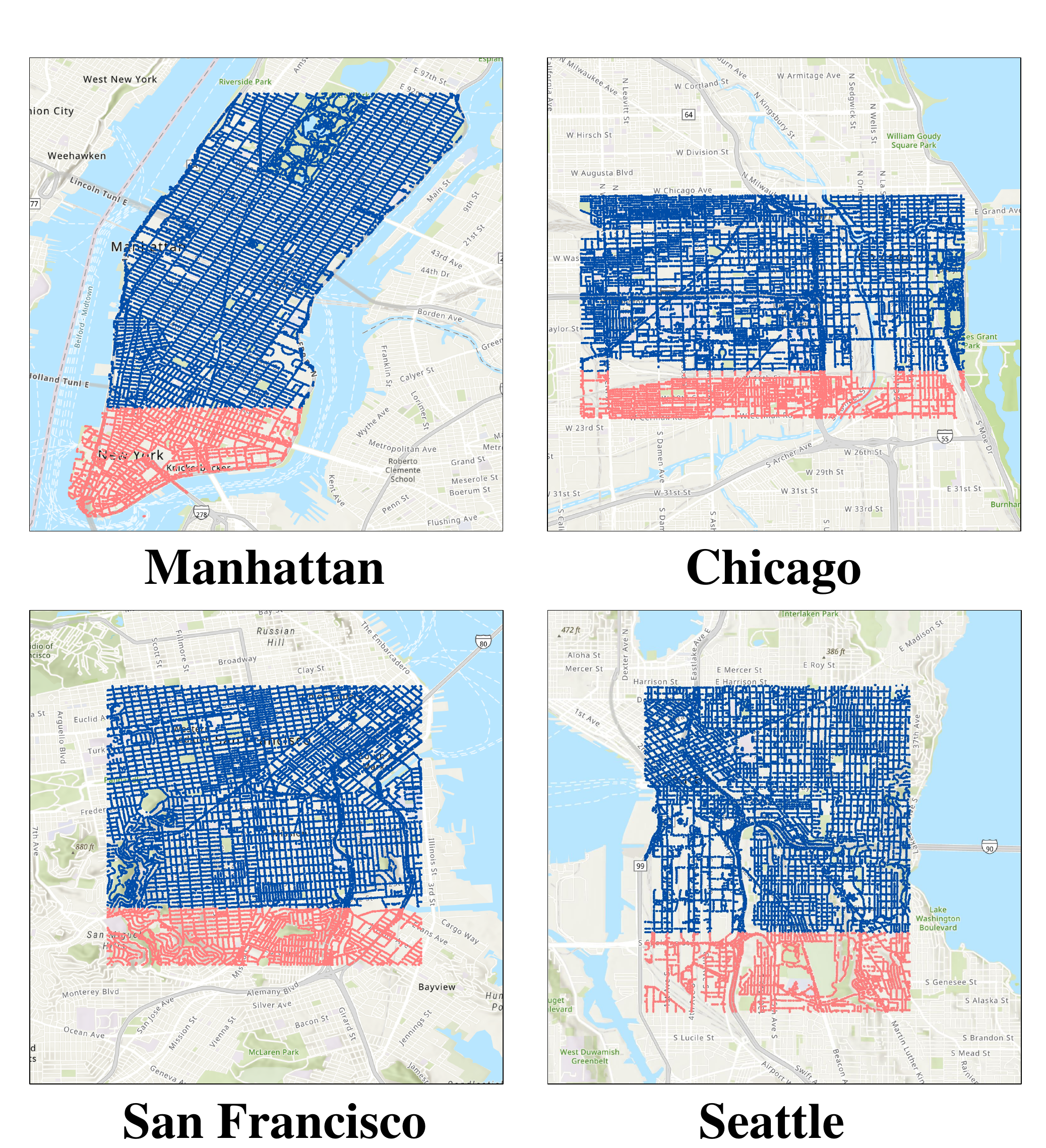}
    \caption{Visualization of training and testing splits for each city in our VIGORv2 dataset. Blue lines indicate the training portion. Red lines indicate the testing portion.}
    \label{fig:test_split}
\end{figure}

\begin{figure*}[!h]
    \centering
    \includegraphics[page=11,width=1.0\textwidth, trim={5cm 0cm 0cm 0cm}, clip]{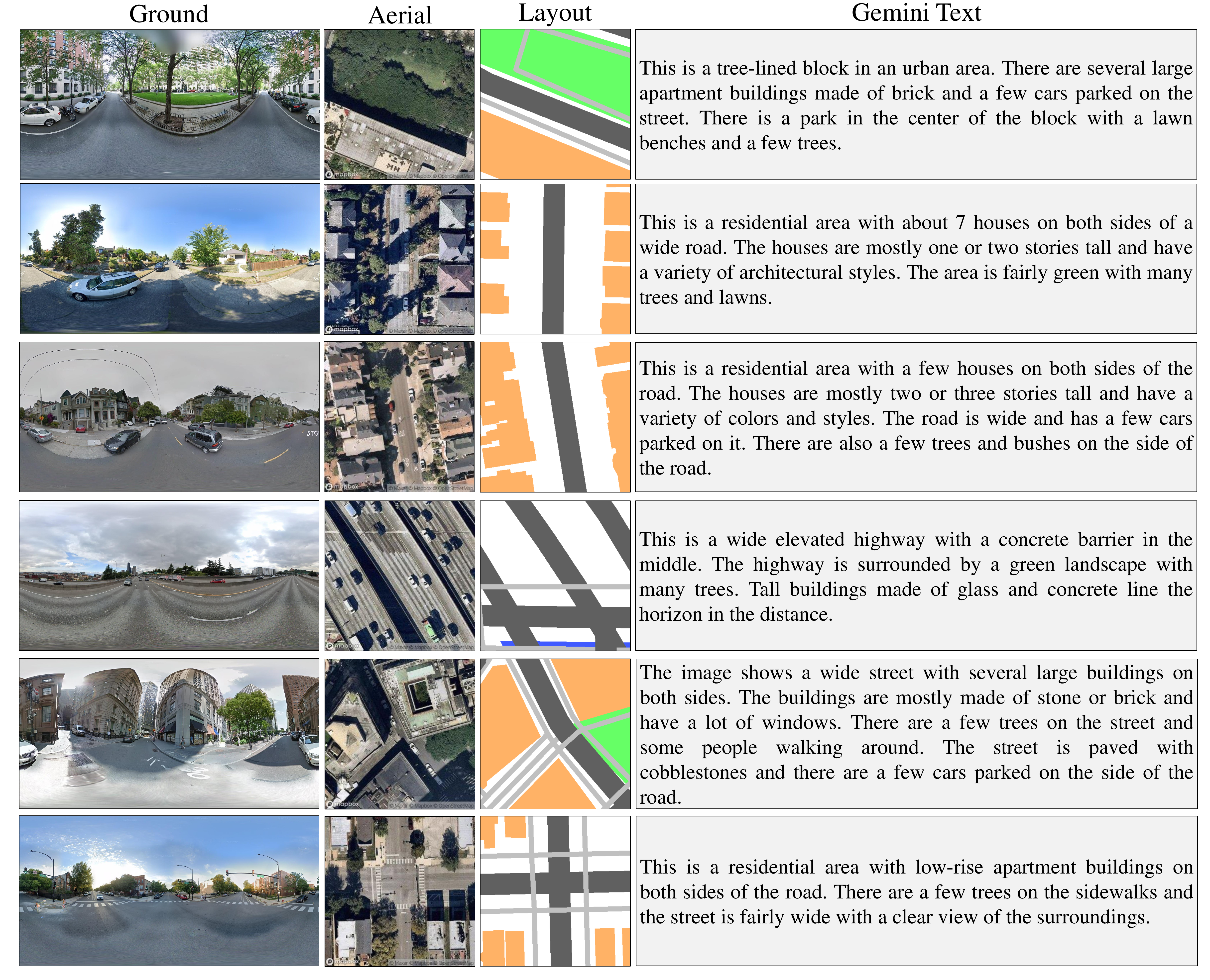}
    \caption{More samples from our proposed VIGORv2 dataset. From left to right are ground images, aerial images, BEV layout maps, and text descriptions for ground images.}
    \label{fig:dataset}
    \vspace{-5pt}
\end{figure*}

\label{sec::dataset}
In our VIGOR\+, we propose a new split that maximizes the discrepancy between training and testing samples by sampling northern and southern regions within the city. We further visualize this split on all four cities in \cref{fig:test_split}. Also, We provide more randomly sampled aerial, ground, text, and BEV layout samples from VIGOR\+ in~\cref{fig:dataset}. These samples demonstrate that our newly proposed dataset, VIGOR\+, contains diverse street layouts in different environments.

\section{Text Data Discussion}
\label{sec::text}
In this section, we will give more detail about the text data used in our \ourmodel{}. The discussion will focus on two main points: the prompt used to collect the Gemini descriptions, and the four types of text conditions used in \ourmodel{} (sec5.4 in the main script)

To collect the ground view descriptions, we passed the following prompt along with the ground image to Gemini:

\textit{You are an AI visual assistant who greatly understands geospatial data. Please generate a paragraph to give a high-level description of the image below with the following constraints. Your output should only be this paragraph without any introductory sentences. Please follow the following rules:
\begin{enumerate}
    \item focus on giving a general description of the place in the image, don't include small-level details like pedestrians and cars.
    \item focus on the buildings, if any, their type, and the number of close-by buildings.
    \item Do not care about any weather conditions, we want a description of the geospatial area
    \item do not include the following words and or any similar words: 'panorama', '360', 'sky', 'car', 'truck', 'pedestrian'
    \item describe the environment type, for example residential, highway, urban, rural .. etc
    \item be limited to about 50 words maximum
    \item if there are people do not pay attention to them and consider them blurred out
    \item please only generate the output directly, do not add any introductory sentences, and only output the description paragraph
    \item If you have any limitations do not mention them, never talk about them
    \item  Only output in English
\end{enumerate}}

\noindent The constraints were both technical to control Gemini's output and minimize hallucinations, and logical, to give accurate descriptions of the ground image, e.g., points 1, 2, 3, and 5. In point 6, we limit the number of words in the output to have consistent samples. As Gemini is designed for chatbot applications, its output is often conversational with redundant greetings and introductory words. We consider these outputs as noise and ask not to include them. Lastly, we make sure to only output in English as we noticed that sometimes, for some reason, Gemini outputs Japanese words. 

After collecting the ground image text descriptions utilizing the prompt above, we thoroughly analyzed our \ourmodel{} on four different cases of text conditions. The constant text prompt is a generic prompt that describes any aerial image, and the same condition is used for all training samples. The constant text prompt was as follows: "\textit{Realistic aerial satellite top view image with high-quality details with buildings and roads}". 

The city text prompt adds a dynamic variable that indicates which city the corresponding sample belongs to. We believe that this addition helps guide the model to differentiate between cities, thus, generating better and diverse samples. This prompt was defined as follows: "\textit{Realistic \{\textbf{CITY}\} aerial satellite top view image with high-quality details with buildings and roads in \{\textbf{CITY}\}}". 

The raw text prompt directly conditions Gemini's output descriptions. An example of such a prompt can be seen in ~\cref{fig:dataset}. These descriptions carry excessive details about the ground image, leading to noise synthesis as shown in Tab. 4 in the main script. Thus, we developed our dynamic text which extracts only relevant information from the collected Gemini descriptions. This relevant information was represented in the form of key phrases. The dynamic text prompt utilized both dynamic city assignment and key-phrase extraction, and was structured as follows: "\textit{Realistic \textbf{CITY} aerial satellite top view image with high-quality details with buildings and roads in \textbf{CITY} that probably has the following objects and characteristics: \textbf{KEYWORDS}}". Examples of dynamic texts are shown in \cref{fig:same_area,fig:cross_area}.

To extract the key phrases from the raw Gemini output, we adopt the Maximal Marginal Relevance (MMR) technique~\cite{MMR}. MMR can be used to rank key phrases based on their relevance to the text content, while also considering their diversity. 
The ranking score $M$ is given as follows,
\begin{equation}
\label{eq:MMR}
   M  \stackrel{\text{def}} {=} \underset{D_i\in S}{argmax} [ \lambda(\Psi(D_i, Q)) - (1-\lambda) \underset{D_j \in \mathcal{S}} {max} (\Psi (D_i, D_j)) ],
\end{equation}
where $Q$ is the Gemini query text, $D_i$ denotes the key phrase to be ranked, $D_j$ are all other remaining key phrases in the document (excluding $D_i$), $\mathcal{S}$ is the set of all ranked key phrases, $\Psi(\cdot,\cdot)$ stands for the cosine similarity between two phrases. This equation controls the trade-off between the relevance and diversity of the extracted key phrases by the $\lambda$ parameter. 
In our experiments, we empirically set the diversity parameter $\lambda = 0.3$, $m=5$, and $N=3$, which yielded the best trade-off between diversity and relevance. 

\section{GeoDTR+ augmentation}
\label{secc::app1}

As mentioned in the manuscript, we also applied the data augmentation techniques in a more recent cross-view geo-localization model, GeoDTR+~\cite{geodtr+}. The settings are the same as described in Sec. 6.1 of the manuscript. \Cref{tab:app_samearea_geodtr} summarizes the results. Similar to the conclusion in the manuscript, we also observed cross-area performance improvement on GeoDTR+ with our data augmentation. For example, R@1 increases from $63.65\%$ to $65.71\%$ on cross-area while $p_o = 0.6$. However, we did not observe the performance increase in the same-area protocol. This might be the better baseline performance of GeoDTR+~\cite{geodtr+} than SAFA~\cite{SAFA}. So that the data augmentation cannot significantly improve the same-area performance.

\begin{table}[!ht]
    \centering
    \footnotesize
    \setlength{\tabcolsep}{3.3 mm}
    \begin{tabular}{cc|cccc}
    \toprule  
     & $p_\circ$ & R@1 $\uparrow$ & R@5$\uparrow$ & R@10$\uparrow$ & R@1\%$\uparrow$ \\
    \toprule
    \multirow{4}{*}{\rotatebox{90}{same-area}} & 0 & 90.21\%         & 96.39\%          & 97.51\%          & 99.47\% \\
      & 0.4 & 90.05\%          & 96.30\%          & 97.48\%          & \textbf{99.61}\% \\
    & 0.6 & \textbf{90.50}\% & \textbf{96.46}\% & \textbf{97.57}\% & 99.54\% \\
    & 0.8 & 90.11\% & 96.34\% & 97.33\% & 99.45\% \\ \midrule
    \multirow{4}{*}{\rotatebox{90}{cross-area}} & 0 & 63.65\%        & 81.01\%         & 86.46\%         & 96.85\% \\
     & 0.4 & 65.48\%         & 82.16\%         & 87.39\%         & 97.03\% \\
    & 0.6 & \textbf{65.71}\% & \textbf{82.53}\% & \textbf{87.80}\% & \textbf{97.13}\% \\
    & 0.8 & 64.54\% & 81.39\% & 87.15\% & 96.93\% \\
    \bottomrule
    \end{tabular}
    \caption{Results of our data augmentation on SAFA. $p_\circ=0$ indicates no augmentation is applied.}
    \label{tab:app_samearea_geodtr}
\end{table}

\section{Sketch-based Region Search Discussion}
\begin{figure*}[!h]
    \centering
    \includegraphics[page=17,width=1.0\textwidth, trim={0cm 2cm 15cm 5cm}, clip]{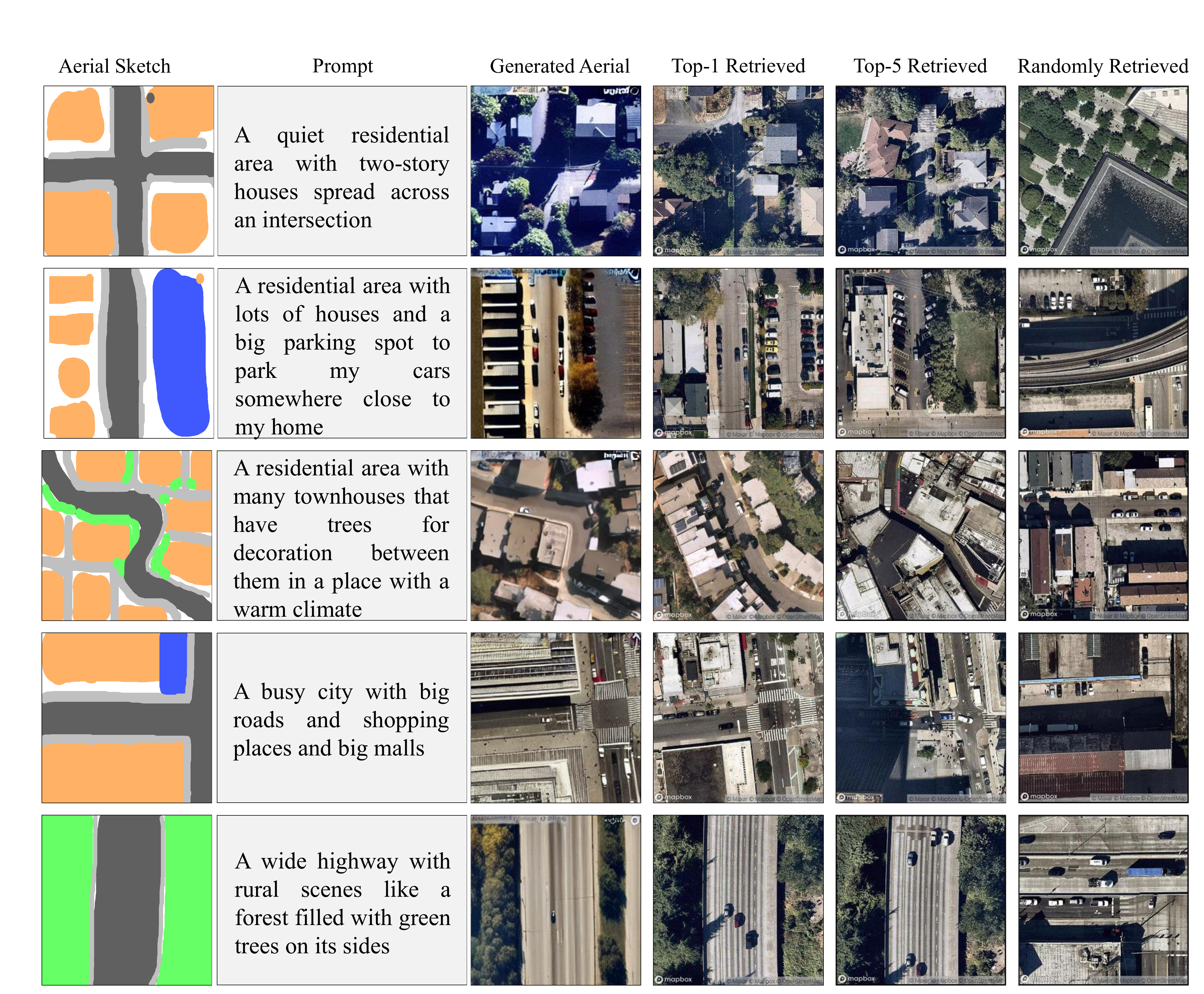}
    \caption{More retrieval samples, and the survey images shown to the participants. We analyzed five groups of images in three cases, top-1, top-5, and a randomly selected image}
    \label{fig:app2_samples}
    \vspace{-10pt}
\end{figure*}

This section presents a more detailed discussion of the sketch-based region search application presented in Sec. 6.2 of the main manuscript. We show additional samples, including an intersection and a swirly road, as shown in \cref{fig:app2_samples}. It is clear how the top retrieved images represent the given sketch and text. Although the street layout does not exactly match the third example, the environment is similar, a residential area in a warm climate. This discrepancy may result from the limited pool of referencing aerial images, where an exact match for the swirly road may not exist, but a close one does.

\begin{figure*}[!h]
    \centering
    \includegraphics[page=18,width=1.0\textwidth, trim={0cm 2cm 5cm 5cm}, clip]{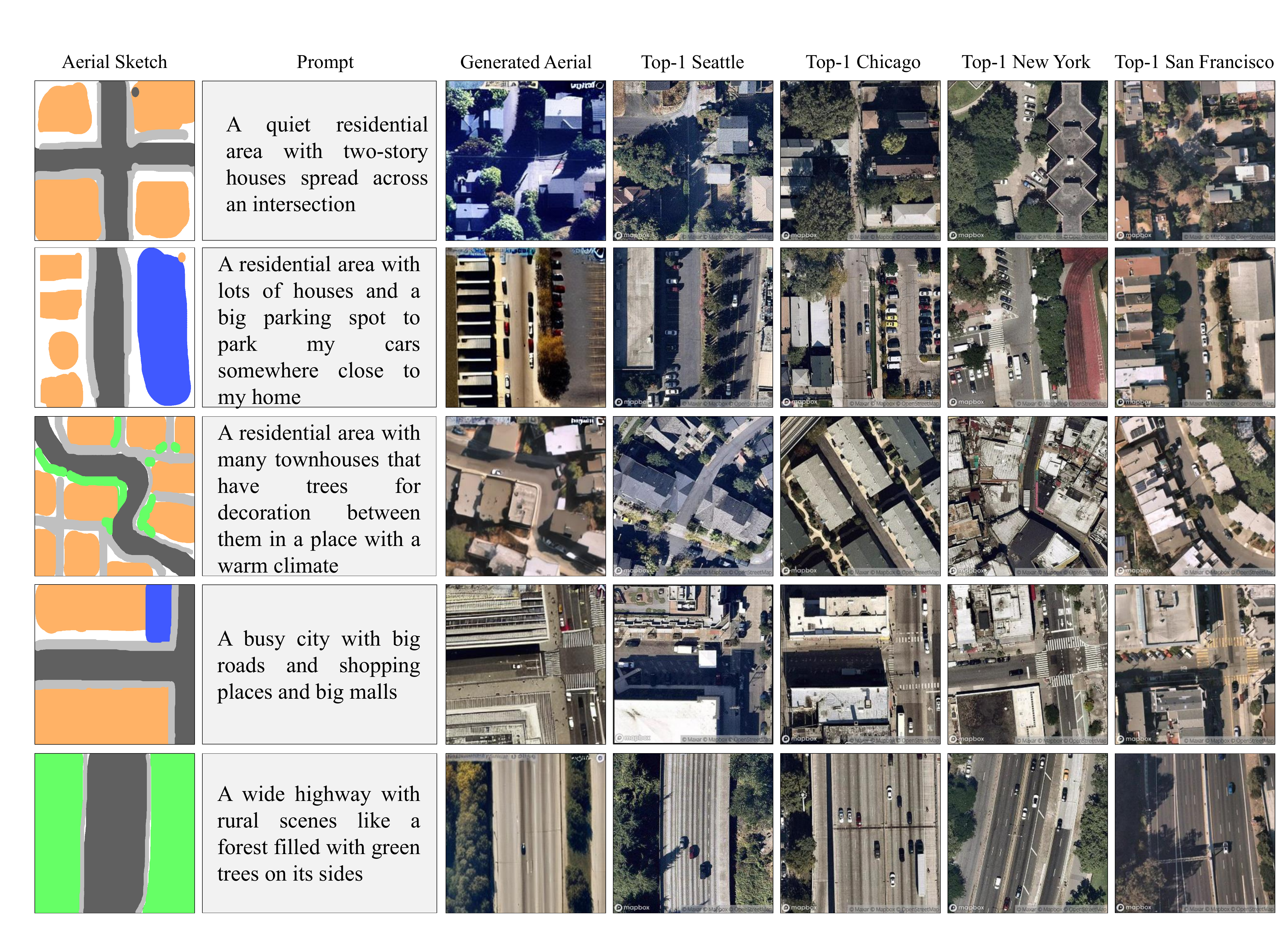}
    \caption{More cross-city results for the sketch-based region search application. Most images reflect both the sketch and text, even across different cities}
    \label{fig:app2_cities}
    \vspace{-10pt}
\end{figure*}

We observed consistent retrieval results even with a larger aerial database. For example, we tried retrieving aerial images from the four cities of VIGOR\+, as shown in \cref{fig:app2_cities}. Overall, the retrieved images align well with the given sketch and text in all cities. Any inconsistencies, if present, would appear in the layout rather than the environment. For instance, all images in the last row represent a highway, a parking lot exists in the second-row images, and an intersection appears to the right of the fourth-row images. These results demonstrate \ourmodel{}'s robustness and how it is applicable in cross-domain applications. 

As mentioned in the main manuscript, we conducted a survey with 61 participants to quantitatively evaluate the retrieval results of our pipeline. The images shown to each volunteer are illustrated in \cref{fig:app2_samples}. Participants rated how well the retrieved image corresponded to the given sketch and text pairs on a scale from 1 to 10. Ratings above 5 were considered agreements that the image matched the sketch and text pair, while ratings below 5 were considered disagreements. With this convention, 66\%, 60\%, and 24\% of the volunteers agreed that the top-1, top-5, and randomly selected images, respectively, represent the given sketch and text. 

\label{secc:app2}

\section{Limitations}
\label{sec::limitation}
In this paper, we paved the way for this challenging research, introducing a new dataset as well as new image quality measurements as a foundation for future comparative studies in this emerging field. The challenges of this task arise from significant variations in perspective, occlusions of objects, and the varying range of visibility between aerial and ground views.
Despite these challenges, our proposed algorithm demonstrates proficiency in preserving geometric information due to the explicit conditioning of the layout maps into our model. One of the limitations of our algorithm is the inability to generate images capturing the location of movable objects like cars and pedestrians. This limitation resulted from the unavailability of synchronized training data, particularly in obtaining a timely synchronized ground-aerial dataset.
Looking ahead, one promising avenue for future exploration involves further enhancing our proposed models for the quality of image synthesis. Additionally, addressing the scarcity of synchronized datasets, future work could focus on the creation of larger and synchronized datasets including diverse cities across continents. This expansion aims to enable the scalability of the proposed methods, advancing more comprehensive and globally applicable systems.

\section{Societal Impact}
\label{sec::social_impact}

In this paper, our novel GPG2A is effective in many areas, as stated in the main script, such as land use classification~\cite{land_use1,land_use2}, urban planning~\cite{urban_planning}, destruction detection~\cite{des_detect}, transportation~\cite{transportation1,transportation2,transportation3} and socioeconomic studies~\cite{socioeco,econmics_ieee}. The predicted BEV layout can also be an auxiliary signal to the positioning system, i.e. comparing the BEV layout to the map to estimate the location. Thus, to this end, the proposed GPG2A will advance the research in both cross-view image synthesis and image geo-localization which will eventually benefit the society. Our proposed VIGORv2 dataset is complementary to the original VIGOR dataset with newly collected center-aligned aerial images, BEV layout maps, and text descriptions for ground images. This dataset will advance future research in this direction and inspire further researchers to work on this problem. Our new aerial image quality evaluation metrics provide a new tool in this topic to help researchers understand the quality of the synthesized images. To this end, this work will benefit the community and advance the research in this area.

\end{document}